\documentclass[10pt,twocolumn,letterpaper]{article}

\PassOptionsToPackage{table}{xcolor}
\usepackage{cvpr}

\usepackage{algorithm}
\usepackage{algpseudocode}

\definecolor{cvprblue}{rgb}{0.21,0.49,0.74}
\usepackage[breaklinks,colorlinks,allcolors=cvprblue,hypertexnames=false]{hyperref}

\title{Generator-Aligned Reward Modeling with Target-Aware Bounded Optimization for Visual Generation}

\author{%
  Ziqiao Weng\textsuperscript{1,2}\thanks{Work done during an internship at TeleAI.},
  Yuanzhi Liang\textsuperscript{1},
  Haibin Huang\textsuperscript{1},
  Weidong Cai\textsuperscript{2},
  Xuelong Li\textsuperscript{1}\\[0.6em]
  \small
  \textsuperscript{1} Institute of Artificial Intelligence, China Telecom (TeleAI)\\
  \small\textsuperscript{2} The University of Sydney\\[0.3em]
}

\hypersetup{
  pdftitle={\thetitle},
  pdfauthor={Ziqiao Weng, Yuanzhi Liang, Haibin Huang, Weidong Cai, Xuelong Li}
}

\begin{document}
\pagestyle{empty}

\maketitle
\thispagestyle{empty}

\begin{abstract}
Reward-guided post-training has substantially improved the generation quality and instruction-following capabilities of visual generative models. However, fine-grained visual-attribute instruction following remains challenging, even when the desired attributes are already within the pretrained generator's capabilities. Addressing this challenge requires reward supervision that captures subtle attribute differences while remaining relevant to the generator's output distribution. Moreover, unconstrained reward maximization may lead to overoptimization and degrade generation quality. We propose GenTA, a Generator-relevant and Target-Anchored reward post-training framework for fine-grained visual-attribute instruction following. GenTA consists of two complementary components. Generator-guided Discriminative Reward Modeling (GDRM) constructs generator-relevant supervision through generator-guided real-example selection and content-matched comparisons, enabling reward models to distinguish fine-grained visual attributes from unrelated content variations. Target-Anchored Reward Optimization (TARO) calibrates a smooth, upper-bounded reward objective using real positive examples, reducing excessive optimization as the target reward level is approached. TARO is optimizer-independent and supports both direct reward-gradient and policy-based post-training. Experiments on image and video generation demonstrate that GenTA improves fine-grained visual-attribute instruction following while maintaining competitive generation quality, highlighting the complementary benefits of generator-relevant reward learning and target-anchored optimization.


\end{abstract}

\section{Introduction}
\label{sec:introduction}

\begin{figure}[t]
    \centering
    \includegraphics[width=\linewidth]{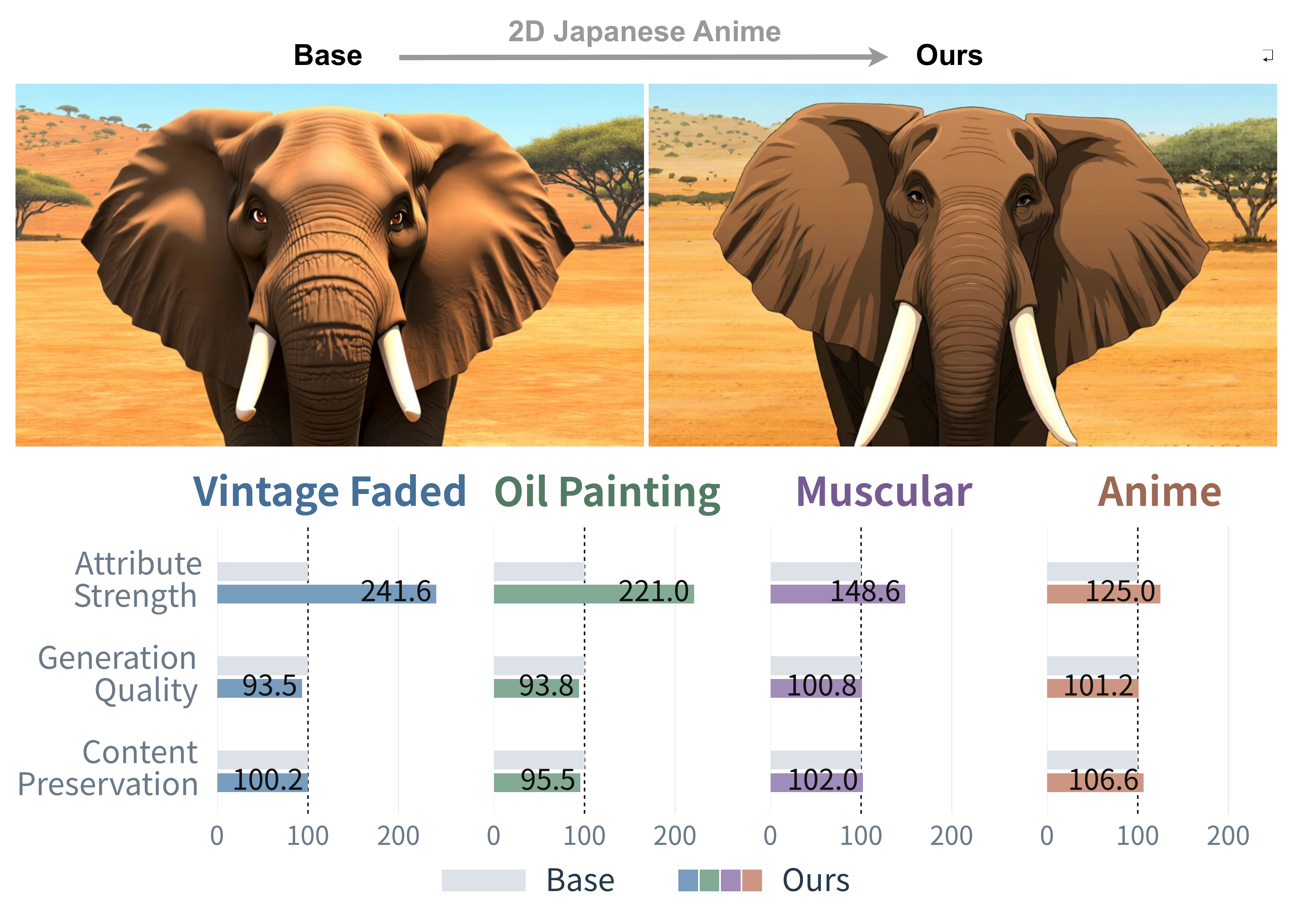}
    \caption{GenTA enhances target attributes while preserving generation quality and content. Top: Base-to-GenTA comparison for 2D Japanese Anime. Bottom: Normalized comparisons (Base = 100, dashed line) across four target attributes—Vintage Faded, Oil Painting, Muscular, and Anime—in terms of attribute strength (20th-percentile attribute score), generation quality (HPSv3 for images and VideoScore2 visual quality for videos), and content preservation (VQAScore for images and VideoScore2 text-to-video alignment for videos).}
    \label{fig:teaser}
\end{figure}

Recent advances in visual generative models have enabled high-quality image and video synthesis with increasingly complex content and motion~\citep{zhang2024vast}. Reward-guided post-training has further improved generation quality and instruction following through differentiable reward feedback and reinforcement learning ~\citep{gao2022overoptimization,black2023ddpo}. Despite this progress, fine-grained visual-attribute instruction following remains challenging. For example, a video generator may produce generic cartoons or three-dimensional animation when prompted for Japanese anime, capturing the broad semantics while missing the requested visual style. Such failures do not necessarily indicate a lack of relevant knowledge. The desired visual attributes may already exist in the pretrained distribution but remain weakly or inconsistently expressed. This motivates post-training that improves the reliable expression of existing visual capabilities while preserving general generation performance.

However, effective fine-grained post-training faces two challenges. First, reward models must distinguish subtle differences in how the target visual attribute is expressed while accounting for the generator's existing capabilities. General-purpose rewards may capture broad visual semantics without reliably distinguishing the requested attribute from visually similar alternatives. Learning from real data alone is also challenging, as positive and negative examples often differ in unrelated content, leading to spurious correlations. Moreover, different generators exhibit different responses to the same instruction. Reward supervision constructed independently of the generator may therefore be poorly aligned with its output distribution. Second, reward optimization must provide appropriate guidance toward the desired outcome. Unconstrained reward maximization may exaggerate the requested visual attribute or exploit reward-model errors, degrading generation quality~\citep{gao2022overoptimization,black2023ddpo}. While reference-based regularization can limit distribution drift, it does not explicitly specify a target level of attribute expression. These challenges call for generator-relevant reward supervision and target-anchored optimization.

To address these challenges, we propose GenTA, a Generator-Relevant and Target-Anchored reward post-training framework for fine-grained visual-attribute instruction following. GenTA integrates Generator-guided Discriminative Reward Modeling (GDRM) and Target-Anchored Reward Optimization (TARO) to jointly improve reward learning and optimization. Specifically, GDRM leverages controlled generations with and without the target condition to identify generator-relevant regions of the real-data distribution. It further constructs content-matched real comparisons to reduce content confounding and improve discrimination of the requested visual attributes. Together, these designs enable more discriminative and generator-relevant reward learning.

Building on GDRM, TARO introduces a target-anchored objective for controlled reward optimization. Rather than maximizing reward scores indefinitely, TARO estimates a target reward level from real positive examples and transforms the learned reward into a smooth, upper-bounded function. This transformation preserves useful optimization signals when the target attribute is under-expressed, while progressively reducing the incentive for further reward increases as the calibrated target is approached. Importantly, TARO is independent of the underlying optimization algorithm and can be integrated with both direct reward-gradient and policy-based post-training.

We evaluate GenTA on fine-grained visual-attribute instruction following in image and video generation, covering targeted visual styles and appearance details. Extensive experiments demonstrate improved target attribute expression while maintaining competitive generation quality, as shown in Fig.~\ref{fig:teaser}. Our main contributions are summarized as follows:

\begin{itemize}
    \item We introduce GenTA, a reward-guided post-training framework that combines generator-relevant reward learning with target-anchored optimization for fine-grained visual-attribute instruction following.

    \item We propose GDRM, which constructs discriminative reward supervision through generator-guided real-example selection and content-matched comparisons, reducing distribution mismatch and content confounding. We further propose TARO, an optimizer-independent, target-calibrated reward objective that moderates excessive reward maximization.


    \item Experiments on image and video generation demonstrate the effectiveness of GenTA across diverse post-training strategies. GDRM consistently enhances fine-grained instruction following while preserving competitive generation quality. TARO further improves the balance between target fidelity and generation quality, mitigating quality degradation during reward-gradient optimization while consistently improving target alignment across different optimization strategies.

\end{itemize}

\section{Related Work}
\label{sec:related_work}

\subsection{Reward Modeling for Visual Generation}
\label{sec:reward_modeling}

Learned reward models provide supervision for evaluating and improving visual generations. ImageReward, PickScore, HPSv2, and HPSv3 learn image preferences from human comparisons, covering diverse aspects of image quality~\citep{xu2023imagereward,kirstain2023pickapic,wu2023hpsv2,ma2025hpsv3}. For video generation, VideoScore, VisionReward, and VideoReward incorporate temporal and semantic dimensions~\citep{he2024videoscore,xu2024visionreward,liu2025videoreward}, while VideoScore2 jointly assesses quality, alignment, and consistency~\citep{he2025videoscore2}. Latent reward models further provide feedback throughout denoising~\citep{mi2025latentreward}. Recent style-specific models include StyleDoctor, which evaluates style consistency using a multimodal generative model, and Style-GRPO, which uses VLM-based scoring for style fidelity and content preservation~\citep{he2026styledoctor,zhao2026stylegrpo}. Unlike these VLM-based architectures, our method employs lightweight, attribute-specific reward models operating directly on generator latents.

GDRM complements reward modeling by constructing reliable real-data supervision aligned with a specific generator. Real attribute labels can be confounded by subjects and scenes, while real examples may poorly represent the generator's output distribution~\citep{liang2019vrr}. GDRM addresses both challenges by organizing real examples according to the generator's empirical output geometry and retrieving content-matched contrasts with opposing attribute labels, yielding generator-relevant and attribute-discriminative supervision.

\subsection{Reward-Guided Post-Training of Diffusion and Flow Models}
\label{sec:reward_guided_post_training}

Reward-guided post-training includes differentiable reward optimization, policy-gradient methods, and preference-based learning. DRaFT, AlignProp, and VADER propagate reward gradients through sampling to update visual generators~\citep{clark2024draft,prabhudesai2023alignprop,prabhudesai2024vader}. DDPO formulates diffusion denoising as a sequential decision process, while Flow-GRPO adapts online reinforcement learning to flow-matching models~\citep{black2023ddpo,liu2025flowgrpo}. DanceGRPO extends group-relative optimization to diffusion and rectified-flow generators, and ViPO introduces spatially and temporally aware advantage estimation~\citep{xue2025dancegrpo,ni2026seeing}. Alternatively, Diffusion-DPO, D3PO, and Flow-DPO optimize directly from pairwise preferences without separately trained reward models~\citep{wallace2024diffusiondpo,yang2024d3po,liu2025videoreward}.

Recent studies address reward reliability and overoptimization in visual generation~\citep{liu2026learning,li2026rethinking}. TDPO-R tackles temporal inductive bias and critic primacy bias in diffusion policy optimization~\citep{zhang2024tdpo}, while GRPO-Guard mitigates flow-matching overoptimization through importance-ratio regulation and timestep-wise gradient reweighting~\citep{wang2025grpoguard}. Focus-N-Fix uses region-aware fine-tuning to constrain changes outside targeted image regions~\citep{xing2025focusnfix}. Unlike spatial preservation, our work targets attribute-conditioned instruction following while preserving generator behavior under non-target conditions. Complementary system-level studies integrate supervised adaptation, reinforcement learning, and preference refinement into broader generation pipelines~\citep{xue2026posttrain,liang2026teleboost,an2026aiflow,shao2026aiflownetworkedge,yuan2026enhancing}. Rather than primarily controlling optimizer updates, TARO calibrates attribute-specific reward targets from real positive examples and applies a smooth, upper-bounded transformation before optimization. These differences in task formulation, conditioning, and reward architecture make StyleDoctor, Style-GRPO, and Focus-N-Fix related but not directly comparable experimental baselines.

\section{Method}
\label{sec:method}
\begin{figure*}[t]
    \centering
    \includegraphics[width=0.9\linewidth]{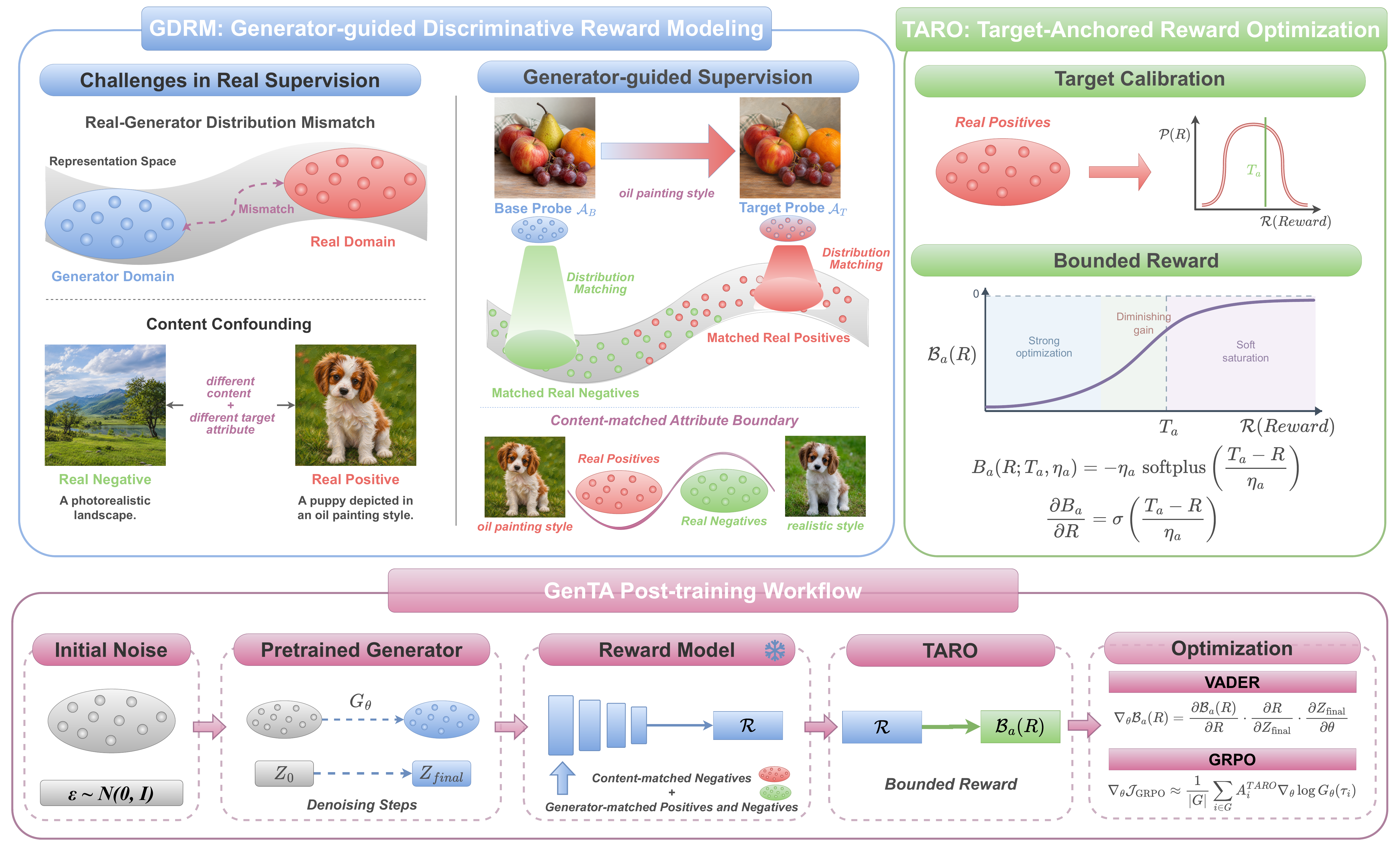}
    \caption{\textbf{GenTA Framework overview.} Using oil-painting style as the target attribute, GDRM leverages paired base/target generator probes and content-matched real negatives to train a generator-aligned latent reward model. TARO calibrates a target reward level from real positives and transforms RM predictions into a bounded reward. During post-training, the bounded reward guides VADER/GRPO optimization to enhance the target attribute while preserving generative capabilities.}
    \label{fig:overview}
\end{figure*}
\subsection{Problem Formulation and Overview}
\label{sec:problem_overview}

Our goal is to improve a pretrained visual generator's ability to follow instructions specifying target visual attributes, using real-data supervision while preserving prompt fidelity, content structure, and general generation quality. Let $G_{\theta_0}$ be a pretrained image or video generator, $a$ a target attribute, and $\mathcal D_a^{\mathrm{real}}=\{(x_j,y_j)\}_{j=1}^{N}$ real supervision with binary labels $y_j\in\{0,1\}$. We train a latent reward model $R_\phi^a$ on real-sample VAE latents $\mathbf z_{\mathrm{VAE}}(x)=E_{\mathrm{VAE}}(x)$. During post-training, it scores the generated final denoised latent $\mathbf z_\theta^{\mathrm{final}}$ in the same latent space. Fig.~\ref{fig:overview} illustrates how GDRM constructs reward supervision, TARO transforms the learned reward, and the resulting bounded objective guides post-training optimization of the generator.

\subsection{Generator-Guided Discriminative Reward Modeling}
\label{sec:garm}

\paragraph{Generator-native Support Calibration.}
A pretrained generator can already express a requested attribute to varying degrees. We probe this capability using paired Base and Target generations that share the content prompt $c_i$, seed $\epsilon_i$, checkpoint, and sampling settings, changing only the attribute condition:
\begin{equation}
    \small
    x_i^B=G_{\theta_0}(c_i;\epsilon_i),\qquad
    x_i^T=G_{\theta_0}(\mathcal{T}_a(c_i);\epsilon_i),
\end{equation}
where $\mathcal{T}_a$ inserts the attribute condition. The resulting anchor sets $\mathcal A_B$ and $\mathcal A_T$ characterize empirical Base and Target supports in a common modality-specific descriptor space. Real examples use VAE-encoded latents, while generated anchors use final denoised latents. Let $\widehat{\mathbf h}(\cdot)$ denote their normalized descriptors and $\ell(v,u)=1-\widehat{\mathbf h}(v)^\top\widehat{\mathbf h}(u)$ their cosine distance. For $s\in\{B,T\}$, the mean distance to the $k$ nearest anchors is:
\begin{equation}
    \small
    d_s(x)=\frac{1}{k}\sum_{u\in\operatorname{kNN}(x;\mathcal A_s)}\ell(x,u),
    \label{eq:support_distance}
\end{equation}
where $\operatorname{kNN}$ selects neighbors in the descriptor space. This distance measures a real sample's proximity to each generator support. Since Base and Target anchors may have different local densities, we calibrate their support radii independently using leave-one-out anchor neighborhoods:
\begin{equation}
    \small
    r_s(u_i)=\frac{1}{k}\sum_{u\in\operatorname{kNN}(u_i;\mathcal A_s\setminus\{u_i\})}\ell(u_i,u).
    \label{eq:support_loo}
\end{equation}
The $q$-quantile of these radii defines each support threshold and membership gate:
\begingroup
\setlength{\abovedisplayskip}{4pt}
\setlength{\belowdisplayskip}{4pt}
\begin{equation}
    \small
    \delta_s=Q_q(\{r_s(u_i)\}_{u_i\in\mathcal A_s}),\qquad
    \Gamma_s(x)=\mathbf{1}[d_s(x)\leq\delta_s].
    \label{eq:support_gate}
\end{equation}
\endgroup
Independent thresholds account for differences in support geometry; a shared radius could be too permissive for a compact support or too restrictive for a dispersed one. The gates estimate empirical generator support and organize real examples according to their relevance to the pretrained generator's output distribution, without changing their original attribute labels. Additional implementation details are provided in the supplementary material.

\paragraph{Multimodal Content-matched Real Attribute Boundary.}
Support matching aligns real supervision with the generator, but positive and negative examples may still differ in subject, scene, appearance, or motion beyond the target visual attribute. For each real positive $p$, we retrieve real negatives $n$ that are similar in non-target content while carrying the opposite attribute label. Semantic similarity $S_{\mathrm{sem}}$ captures factors such as subject and scene, excluding semantic information directly associated with the target attribute; visual similarity $S_{\mathrm{vis}}$ adds cues about appearance and spatial structure. We map both scores to empirical CDF quantiles $Q_{\mathrm{sem}}$ and $Q_{\mathrm{vis}}$ within each positive/negative source-role pair to account for their different scales and distributions. The fused similarity is:
\begin{equation}
    \small
    S_{\mathrm{fused}}(p,n)
    =Q_{\mathrm{sem}}(p,n)^w Q_{\mathrm{vis}}(p,n)^{1-w},
    \qquad w\in[0,1].
\end{equation}
The geometric fusion favors negatives that are jointly similar in semantic content and visual appearance. These opposite-label contrasts reduce reliance on incidental content cues and make differences in the target visual attribute more informative near the local decision boundary. In our experiments, the semantic branch uses internally constructed image and video tags, excluding information associated with the target attribute while retaining other semantic content cues. Further details on the scoring and retrieval procedure are provided in the supplementary material.

\paragraph{Generator-guided Discriminative Supervision and Reward Learning.}
Let $\mathcal P_a$ and $\mathcal N_a$ denote the real positive and negative sets. We compare each sample's normalized distance to the two supports using:
\begingroup
\setlength{\abovedisplayskip}{4pt}
\setlength{\belowdisplayskip}{3pt}
\begin{equation}
    \small
    \rho_s(x)=d_s(x)/\delta_s,\qquad m(x)=\rho_B(x)-\rho_T(x).
    \label{eq:support_margin}
\end{equation}
\endgroup
Here, $\rho_s(x)$ is the distance relative to the calibrated scale of support $s$, and $m(x)\leq0$ indicates that the sample is at least as close to Base as to Target in normalized distance. \Cref{alg:garm-pools} uses the support gates, relative margin, and content matching to construct five training pools.

\begin{algorithm}[t]
\caption{Construction of the five GDRM training pools}
\label{alg:garm-pools}
\begin{algorithmic}[1]
\Require Real sets $\mathcal P_a,\mathcal N_a$, support gates $\Gamma_B,\Gamma_T$, and margin $m$
\State $\mathcal P_{\mathrm{match}}\gets\{x\in\mathcal P_a:\Gamma_T(x)=1\}$
\State $\mathcal P_{\mathrm{cov}}\gets\mathcal P_a\setminus\mathcal P_{\mathrm{match}}$
\State $\mathcal N_{\mathrm{gray}}\gets\{x\in\mathcal N_a:\Gamma_T(x)=1,\ \Gamma_B(x)=0\}$
\State $\mathcal N_{\mathrm{elig}}\gets\mathcal N_a\setminus\mathcal N_{\mathrm{gray}}$
\State Retrieve $\mathcal N_{\mathrm{hard}}\subseteq\mathcal N_{\mathrm{elig}}$ by content matching to positives
\State $\mathcal N_{\mathrm{rem}}\gets\mathcal N_{\mathrm{elig}}\setminus\mathcal N_{\mathrm{hard}}$
\State $\mathcal N_{\mathrm{match}}\gets\{x\in\mathcal N_{\mathrm{rem}}:\Gamma_B(x)=1,$
\Statex \hspace{\algorithmicindent}$[\Gamma_T(x)=0\ \lor\ m(x)\leq 0]\}$
\State $\mathcal N_{\mathrm{cov}}\gets\mathcal N_{\mathrm{rem}}\setminus\mathcal N_{\mathrm{match}}$
\State \Return $\mathcal P_{\mathrm{match}},\mathcal P_{\mathrm{cov}},\mathcal N_{\mathrm{hard}},\mathcal N_{\mathrm{match}},\mathcal N_{\mathrm{cov}}$
\end{algorithmic}
\end{algorithm}

All Target-supported positives are retained, including Base-preferred boundary positives, which provide useful target-attribute evidence within overlapping supports. The remaining positives form the coverage-positive pool. Negative construction excludes Target-only gray negatives, retrieves content-matched hard negatives from the eligible set, and divides the remaining negatives into Base-supported matched negatives and coverage negatives. Together, the five pools balance generator relevance, local discrimination of the target visual attribute, and broader real-data coverage without changing the original binary labels. We train the latent reward model with binary cross-entropy on real-sample VAE latents and freeze it for target calibration and post-training.

\subsection{Target-Anchored Reward Optimization}
\label{sec:tabo}

Rather than maximizing raw reward indefinitely, TARO reduces the marginal benefit of reward increases as the predicted expression of the target attribute approaches a target level calibrated from real positive examples. For raw reward logit $R$, it defines the smooth, upper-bounded objective:
\begin{equation}
    \small
    \begin{aligned}
    B_a(R;T_a,\eta_a)&=-\eta_a\,\operatorname{softplus}
       \left(\frac{T_a-R}{\eta_a}\right),\\
    \frac{\partial B_a}{\partial R}
      &=\sigma\left(\frac{T_a-R}{\eta_a}\right).
    \end{aligned}
\end{equation}
Here, $T_a$ is the target level calibrated from real positive examples and $\eta_a>0$ controls the transition width; their calibration is detailed in the supplementary material. The transformed reward is upper-bounded by zero. Its slope is close to one below $T_a$, equals one half at $T_a$, and approaches zero above it. Thus, TARO preserves a strong optimization signal when the target attribute is under-expressed, while smoothly reducing the benefit of further reward increases as the calibrated target is approached and exceeded. The target is a soft saturation point, not a hard clipping threshold. Since TARO transforms the scalar reward before the optimizer acts, it supports both direct reward-gradient and policy-based updates.

\subsection{Optimizer-Specific Realizations}
\label{sec:optimizer_realizations}

\paragraph{Direct Reward-gradient Optimization.}
For differentiable reward optimization, $\partial B_a/\partial R$ scales the reward gradient through the final denoised latent. The remaining gradient path follows the VADER update~\citep{prabhudesai2024vader}.

\paragraph{Advantage-level Adaptation for Group-relative Policy Optimization.}
Vanilla GRPO normalizes rewards by the standard deviation within each prompt group. When applied to TARO-transformed rewards, this normalization can weaken the intended attenuation: as bounded reward differences shrink near the target, the within-group standard deviation may shrink with them. To retain target-dependent attenuation after group-relative normalization, we use a shared scale across groups within each update interval:
\begin{equation}
    \small
    \label{eq:ta_advantage}
    A_i^{\mathrm{adapt}}
      =\frac{\widetilde R_i-\overline{\widetilde R}_G}{v_k+\epsilon},
    \qquad
    \widetilde R_i=B_a(R_i;T_a,\eta_a).
\end{equation}
Here, $R_i$ is the reward logit of sample $i$, and $T_a$ remains fixed after calibration on real positive examples. The shared scale $v_k$ is initialized from a calibration rollout and periodically refreshed using recent group-level bounded-reward dispersion. Smoothed updates and an initial-scale lower bound stabilize the scale, which remains fixed within each update interval rather than being computed from the current group standard deviation. This adaptation is designed to preserve TARO's target-dependent attenuation in group-relative policy optimization. For video optimization, timestep-balanced credit assignment further balances contributions across denoising steps; its weights apply only to the policy optimization term and leave the reference-KL contribution unchanged. Further implementation details are provided in the supplementary material.

\paragraph{Condition-dependent Reference Preservation.}
Reference preservation complements TARO by constraining deviations from pretrained behavior, while TARO controls reward-induced optimization pressure. We combine the task loss with separate KL penalties for target-conditioned and neutral prompts:
\begin{equation}
    \small
    \mathcal L_{\mathrm{total}}=\mathcal L_{\mathrm{task}}^{T}
    +\lambda_T\mathcal L_{\mathrm{KL}}^{T}
    +\lambda_N\mathcal L_{\mathrm{KL}}^{N}.
\end{equation}
The target-conditioned branch receives the attribute reward and is regularized against a frozen checkpoint from an earlier training step. The neutral branch receives no attribute reward and is regularized against the original pretrained generator. The prediction step and regularization coefficients are specified in \cref{sec:setup_post_training}.

\section{Experimental Setup}
\label{sec:experimental_setup}

\subsection{Tasks, Models, and Data}
\label{sec:tasks_models_data}

We study Vintage Faded, Oil Painting, and Muscular image generation with Z-Image, and 2D Japanese anime generation in video. For video post-training, we use Wan2.1-T2V-14B for VADER-based optimization and Wan2.1-T2V-1.3B for Flow-GRPO-based optimization.

Reward models are trained on real image and video data. For image reward modeling, we construct approximately 60K image--text pairs from Re-LAION-Caption19M through quality filtering and seeded sampling, including roughly 1K target-positive examples. The processed video corpus contains approximately 5.5M videos with structured labels. For Anime reward modeling, approximately 10K target-positive videos are combined with negative examples sampled from the full negative pool, yielding approximately 100K training examples. These real-data collections are used exclusively for reward-model training and are distinct from the prompts used for generator post-training.

Post-training prompts are drawn from DanceGRPO. Evaluation uses 300 prompts per image attribute and 200 prompts for Anime. For Muscular, prompts are restricted to people and settings suitable for assessing enhanced muscular definition.

\subsection{Reward Models and Evaluation Protocol}
\label{sec:setup_rm_anchors}

The image reward model is a six-layer latent ViT, while the video reward model combines local spatial tokenization with temporal attention over Wan latents. To construct generator supports for GDRM, we generate paired Base and Target samples using identical content prompts, seeds, checkpoints, and sampling settings, changing only the attribute condition. We use 5K pairs each for Vintage Faded, Oil Painting, and Anime, and 2.5K pairs for Muscular. Their final denoised latents define the Base and Target supports for real-sample matching.

Independent RGB evaluators trained on labeled real examples assess target-attribute expression, while latent reward models provide optimization signals during post-training. Target strength is measured by the mean and 20th percentile (P20) of RGB scores mapped to percentiles using Target anchors. Image quality is evaluated with HPSv2, HPSv3, PickScore, VQAScore, and Q-Align quality and aesthetics. Video quality is evaluated using the arithmetic means of three VBench metrics (subject consistency, dynamic degree, and aesthetic quality) and three VideoScore2 metrics (visual quality, text alignment, and physical consistency). Additional validation and evaluation details are provided in the supplementary material.

\subsection{Post-Training Protocol}
\label{sec:setup_post_training}

For GDRM comparisons, we vary only the reward-supervision construction while keeping the generator, reward-model architecture and training budget, optimizer, prompts, sampling settings, and post-training steps fixed. For optimization comparisons, we fix the reward supervision and compare VADER with VADER+TARO, and Flow-GRPO with Flow-GRPO+TARO, within their respective backbones. For each attribute, the TARO target $T_a$ is set to the 75th percentile of real-positive logits, and the transition width $\eta_a$ to half their interquartile range plus a numerical-stability term. For group-relative policy optimization, advantage-level adaptation uses a periodically updated shared advantage scale. Timestep-balancing weights mitigate the dominance of late flow steps without rescaling the reference terms. For condition-dependent reference preservation, target-conditioned prompts receive the attribute task loss with a weight of 0.01 and use a frozen model snapshot from training step $T_{\mathrm{ref}}=10$ for image attributes or $T_{\mathrm{ref}}=40$ for Anime video as the reference. Neutral prompts receive no attribute task loss and use the original pretrained generator as the reference. The corresponding KL penalties compare predictions only at the last generation step, with weights $\lambda_T=0.03$ and $\lambda_N=0.1$, respectively. Here, $T_{\mathrm{ref}}$ denotes a training checkpoint rather than a flow timestep. Further implementation details of the advantage-level adaptation are provided in the supplementary material.

\section{Results}
\label{sec:results}

\subsection{Generator-Guided Discriminative Reward Modeling}
\label{sec:exp_garm}

To evaluate reward-supervision construction for targeted post-training, we compare three strategies while keeping the generator, optimizer, and training budget fixed. \emph{Direct Real} directly uses real target-positive and target-negative examples with their original labels. \emph{Generated-only} uses the generator's Base and Target anchors. \emph{GDRM} is our proposed method. Additional ablations of support matching and content matching
are provided in the supplementary material. The pretrained generator serves as the Base reference.

\begin{table*}[tbp]
\centering
\caption{\textbf{Post-target results under different reward-supervision constructions.} Target strength is measured by percentile statistics calibrated on the RGB Target-val anchors.  Base is shown only as a reference; \textbf{bold} denotes the best post-trained result and underlining denotes the second-best post-trained result for each metric.}
\label{tab:image-post-target}
\footnotesize
\setlength{\tabcolsep}{2.0pt}
\renewcommand{\arraystretch}{0.96}
\begin{tabular*}{\linewidth}{@{\extracolsep{\fill}}lrrrrrrrr@{}}
\toprule
Method & Mean $\uparrow$ & P20 $\uparrow$ & HPSv3 $\uparrow$ & Q-Align Q. $\uparrow$ & Q-Align Aes. $\uparrow$ & HPSv2 $\uparrow$ & PickScore $\uparrow$ & VQA $\uparrow$ \\
\midrule
\multicolumn{9}{c}{\textbf{\textit{Muscular}}} \\
\addlinespace[0.5pt]
\rowcolor{gray!10}
Base (ref.) & 51.838 & 24.835 & 9.469 & 4.826 & 3.881 & 0.307 & 20.582 & 0.706 \\
Direct Real & \underline{57.341} & 29.196 & \underline{9.340} & \underline{4.753} & \underline{3.778} & \underline{0.308} & \underline{20.554} & \underline{0.684} \\
Generated-only & 55.034 & \underline{31.043} & 8.983 & 4.568 & 3.534 & 0.305 & 20.507 & 0.680 \\
GDRM & \textbf{62.862} & \textbf{36.896} & \textbf{9.549} & \textbf{4.818} & \textbf{3.888} & \textbf{0.312} & \textbf{20.673} & \textbf{0.720} \\

\addlinespace[1.5pt]
\midrule
\addlinespace[0.5pt]
\multicolumn{9}{c}{\textbf{\textit{Oil Painting}}} \\
\addlinespace[0.5pt]
\rowcolor{gray!10}
Base (ref.) & 51.282 & 26.436 & 10.818 & 4.416 & 3.798 & 0.299 & 20.235 & 0.730 \\
Direct Real & \underline{74.455} & \underline{55.177} & \underline{9.787} & \underline{3.912} & \underline{3.636} & \underline{0.282} & \underline{19.992} & \underline{0.687} \\
Generated-only & 67.135 & 45.234 & 9.226 & 3.728 & 3.474 & 0.277 & 19.969 & 0.684 \\
GDRM & \textbf{76.102} & \textbf{58.436} & \textbf{10.146} & \textbf{4.043} & \textbf{3.696} & \textbf{0.291} & \textbf{20.074} & \textbf{0.697} \\

\addlinespace[1.5pt]
\midrule
\addlinespace[0.5pt]
\multicolumn{9}{c}{\textbf{\textit{Vintage Faded}}} \\
\addlinespace[0.5pt]
\rowcolor{gray!10}
Base (ref.) & 50.733 & 22.606 & 11.555 & 4.447 & 3.624 & 0.294 & 19.857 & 0.708 \\
Direct Real & \underline{72.862} & \underline{54.527} & 10.108 & 3.883 & 3.377 & \underline{0.288} & 19.949 & \textbf{0.710} \\
Generated-only & 62.153 & 33.596 & \textbf{10.851} & \textbf{4.146} & \underline{3.488} & 0.285 & \textbf{19.987} & 0.671 \\
GDRM & \textbf{74.425} & \textbf{54.624} & \underline{10.805} & \underline{4.056} & \textbf{3.497} & \textbf{0.292} & \underline{19.971} & \underline{0.709} \\
\bottomrule
\end{tabular*}
\end{table*}

\begin{figure}[!tbp]
    \centering
    \includegraphics[width=\linewidth]{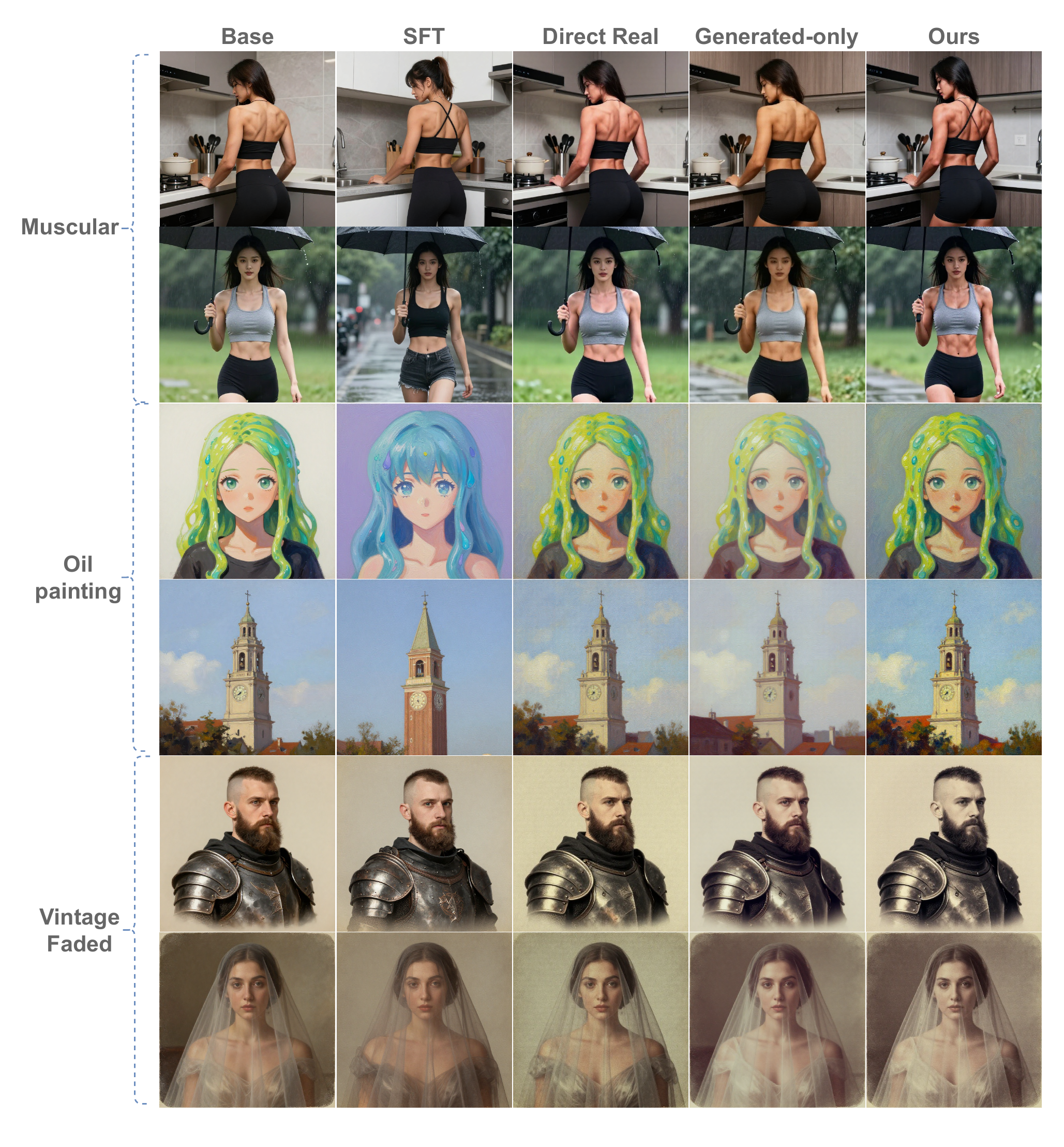}
    \caption{Qualitative comparison of Base, SFT, Direct Real, Generated-only, and GDRM on Muscular, Oil Painting, and Vintage Faded. Examples illustrate target expression and preservation of scene content.}
    \label{fig:image-vis-compare}
\end{figure}

\paragraph{Image Results.}
We evaluate whether GDRM provides effective attribute-specific supervision across diverse image attributes while maintaining general image quality. As shown in \cref{tab:image-post-target}, GDRM achieves the strongest target expression across all three attributes. This improvement is accompanied by consistently strong image-quality scores for Muscular and Oil Painting, while Vintage Faded exhibits a more varied quality trade-off. Overall, the results show that GDRM improves attribute expression, with its effect on general image quality depending on the target attribute. The qualitative comparisons in \cref{fig:image-vis-compare} further corroborate GDRM's effectiveness across all three attributes.

\begin{figure*}[!tbp]
    \centering
    \includegraphics[width=0.95\linewidth]{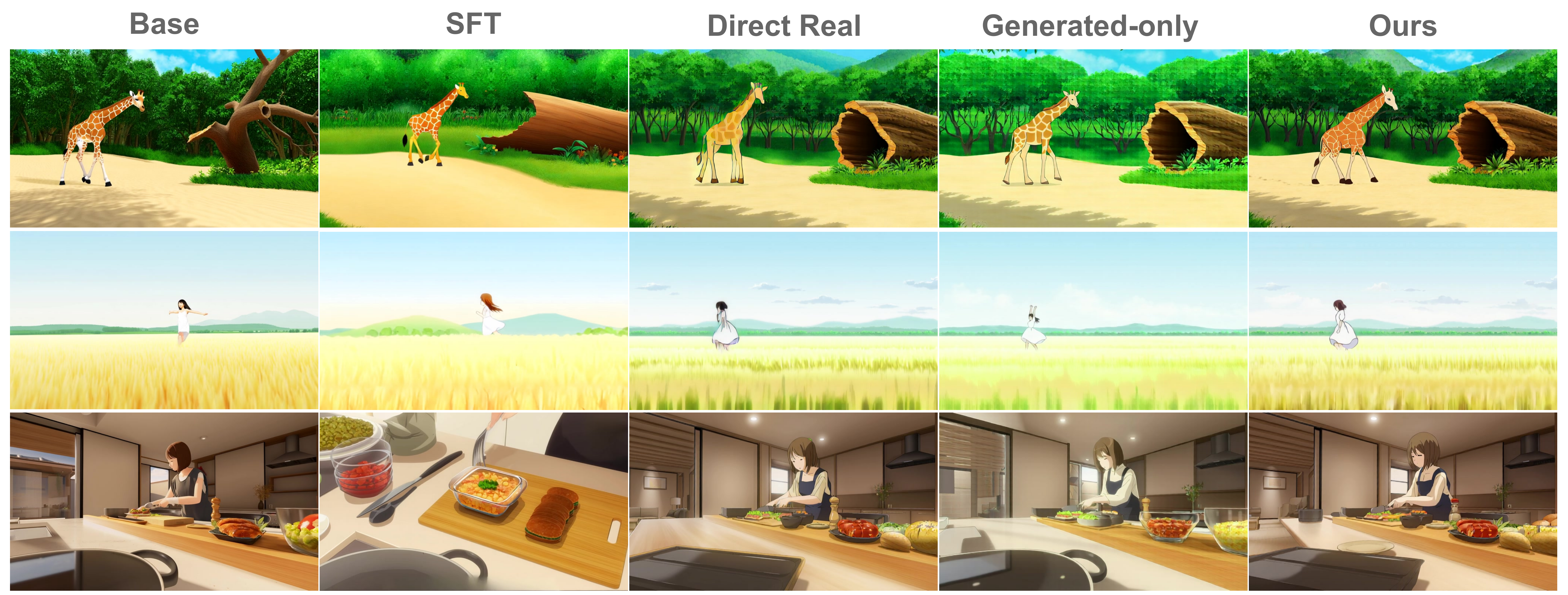}
    \caption{Qualitative comparison of generation methods for 2D Japanese anime animation with Wan2.1-T2V-14B.}
    \label{fig:video-vis-compare}
\end{figure*}

\begin{table*}[t]
\centering
\caption{\textbf{Post-target enhancement on Anime video.} Target strength is measured by RGB target-support percentiles (Mean and P20). VBench metrics include subject consistency (SC), dynamic degree (DD), and aesthetic quality (Aes.); VideoScore2 metrics include visual quality (VQ), text alignment (TA), and physical consistency (PC). Reported means are arithmetic averages of the respective metrics.}
\label{tab:video-post-target}
\footnotesize
\setlength{\tabcolsep}{1.8pt}
\renewcommand{\arraystretch}{0.96}
\begin{tabular*}{\linewidth}{@{\extracolsep{\fill}}lrrrrrrrrrr@{}}
\toprule
 & \multicolumn{2}{c}{Attribute} & \multicolumn{4}{c}{VBench} & \multicolumn{4}{c}{VideoScore2} \\
\cmidrule(lr){2-3}\cmidrule(lr){4-7}\cmidrule(lr){8-11}
Method & Mean $\uparrow$ & P20 $\uparrow$ & SC $\uparrow$ & DD $\uparrow$ & Aes. $\uparrow$ & Mean $\uparrow$ & VQ $\uparrow$ & TA $\uparrow$ & PC $\uparrow$ & Mean $\uparrow$ \\
\midrule
\rowcolor{gray!10}
Base (ref.) & 59.263 & 43.430 & 0.9725 & 0.2117 & 0.6655 & 0.6166 & 3.9720 & 3.8826 & 3.7742 & 3.8763 \\
Direct Real & \underline{68.579} & \textbf{55.417} & \underline{0.9717} & \underline{0.1350} & \underline{0.6362} & \underline{0.5810} & 3.9586 & \textbf{4.1870} & \underline{3.7643} & \underline{3.9700} \\
Generated-only & 68.079 & 53.974 & 0.9684 & 0.1150 & 0.6288 & 0.5707 & \underline{3.9713} & 3.9709 & 3.7016 & 3.8813 \\
GDRM & \textbf{68.722} & \underline{54.267} & \textbf{0.9772} & \textbf{0.1450} & \textbf{0.6434} & \textbf{0.5885} & \textbf{4.0182} & \underline{4.1391} & \textbf{3.8467} & \textbf{4.0013} \\
\bottomrule
\end{tabular*}
\end{table*}


\paragraph{Anime Video Results.}
\begin{figure}[!tbp]
    \centering
    \scalebox{1}[1.2]{%
        \includegraphics[width=\linewidth]{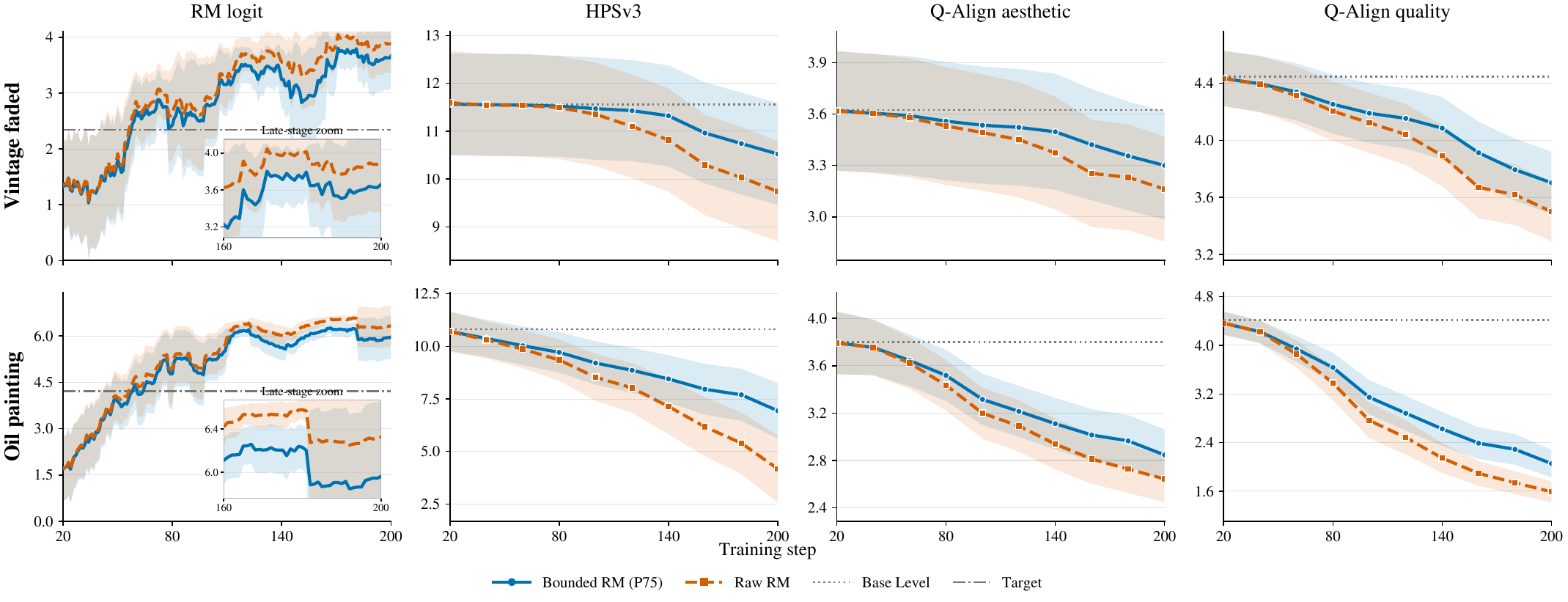}
    }
    \caption{Reward optimization with VADER and VADER+TARO on Vintage Faded (top) and Oil Painting (bottom). Curves show reward logits, HPSv3, and Q-Align aesthetics and quality across training.}
    \label{fig:bounded-vs-raw-rm}
\end{figure}

We further examine whether GDRM generalizes to video reward modeling. GDRM achieves the highest target Mean among the post-trained methods, while Direct Real has a slightly higher P20 (\cref{tab:video-post-target}). GDRM also obtains the best post-trained VBench and VideoScore2 means, indicating a favorable balance between target expression and overall video quality. Although its VBench mean remains below Base, its VideoScore2 mean is higher. The qualitative results in \cref{fig:video-vis-compare} further support the effectiveness of GDRM for video generation.

\subsection{Target-Anchored Reward Optimization}
\label{sec:exp_tabo}

\begin{figure*}[!t]
    \centering
    \includegraphics[width=0.95\linewidth]{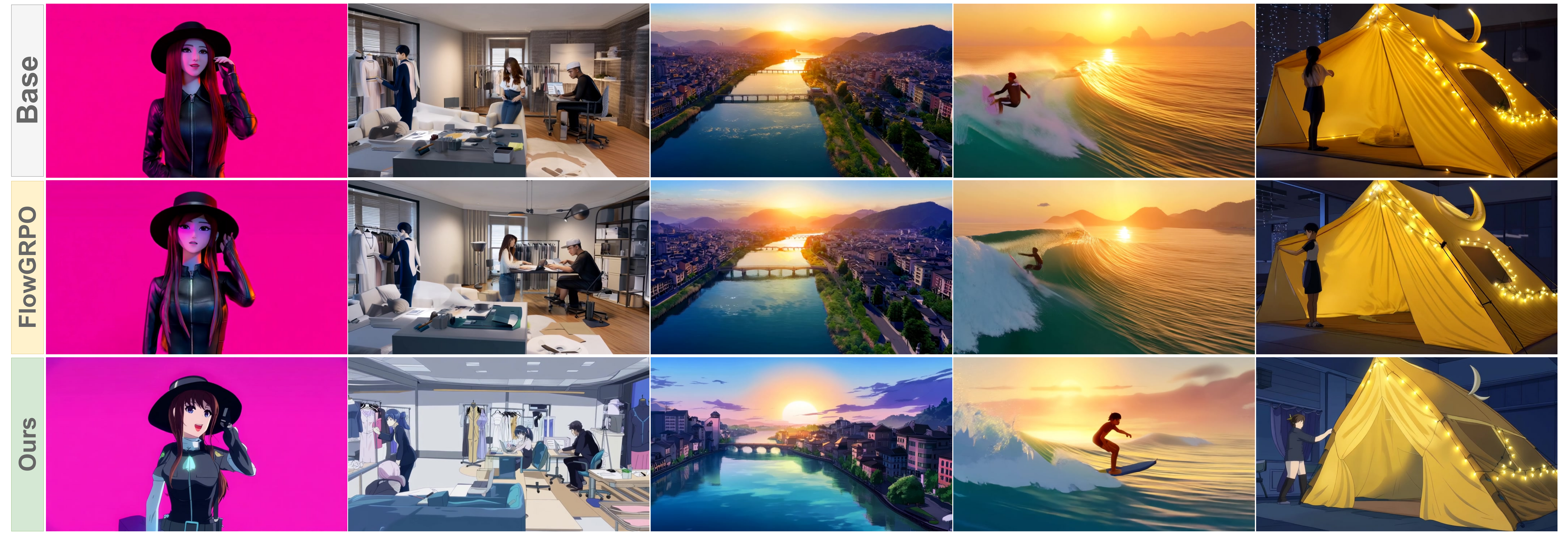}
    \caption{Qualitative comparison of Base, Flow-GRPO, and Flow-GRPO+TARO for 2D Japanese anime animation with Wan2.1-T2V-1.3B.}
    \label{fig:video-vis-GRPO-compare}
\end{figure*}

\begin{table*}[!t]
\centering
\caption{\textbf{Post-training results with and without TARO under two optimization strategies.} Evaluation metrics follow \cref{tab:video-post-target}.}
\label{tab:tabo-ta-grpo-target}
\footnotesize
\setlength{\tabcolsep}{1.2pt}
\renewcommand{\arraystretch}{0.96}
\begin{tabular*}{\linewidth}{@{\extracolsep{\fill}}lrrrrrrrrrr@{}}
\toprule
 & \multicolumn{2}{c}{Attribute} & \multicolumn{4}{c}{VBench} & \multicolumn{4}{c}{VideoScore2} \\
\cmidrule(lr){2-3}\cmidrule(lr){4-7}\cmidrule(lr){8-11}
Method & Mean $\uparrow$ & P20 $\uparrow$ & SC $\uparrow$ & DD $\uparrow$ & Aes. $\uparrow$ & Mean $\uparrow$ & VQ $\uparrow$ & TA $\uparrow$ & PC $\uparrow$ & Mean $\uparrow$ \\
\midrule
\multicolumn{11}{c}{\textbf{\textit{(a) VADER-based optimization}}} \\
\addlinespace[0.5pt]
\rowcolor{gray!10}
Base (ref.) & 59.263 & 43.430 & 0.9725 & 0.2117 & 0.6655 & 0.6166 & 3.9720 & 3.8826 & 3.7742 & 3.8763 \\
VADER & \textbf{69.543} & \textbf{56.695} & 0.9763 & 0.1200 & 0.6359 & 0.5774 & 3.9897 & 3.9715 & 3.7782 & 3.9131 \\
VADER+TARO & 68.722 & 54.267 & \textbf{0.9772} & \textbf{0.1450} & \textbf{0.6434} & \textbf{0.5885} & \textbf{4.0182} & \textbf{4.1391} & \textbf{3.8467} & \textbf{4.0013} \\
\addlinespace[1.5pt]
\midrule
\multicolumn{11}{c}{\textbf{\textit{(b) Flow-GRPO-based optimization}}} \\
\addlinespace[0.5pt]
\rowcolor{gray!10}
Base (ref.) & 61.706 & 45.686 & 0.9790 & 0.1200 & 0.6558 & 0.5849 & 3.9902 & 3.8529 & 3.7866 & 3.8766 \\
Flow-GRPO & 67.227 & 51.002 & \textbf{0.9775} & 0.1450 & \textbf{0.6455} & 0.5893 & \textbf{3.9831} & \textbf{4.0931} & \textbf{3.7686} & \textbf{3.9483} \\
Flow-GRPO+TARO & \textbf{71.684} & \textbf{60.401} & 0.9700 & \textbf{0.2800} & 0.6359 & \textbf{0.6286} & 3.9682 & 4.0812 & 3.7572 & 3.9355 \\
\bottomrule
\end{tabular*}
\end{table*}

To evaluate the effect of TARO on reward optimization while retaining target enhancement, we compare VADER, which performs raw-reward maximization, with VADER+TARO on Wan2.1-T2V-14B. As shown in \cref{tab:tabo-ta-grpo-target}, VADER achieves slightly stronger target expression, whereas VADER+TARO obtains higher VBench and VideoScore2 means. This indicates that TARO reduces the quality cost of reward optimization while preserving substantial attribute enhancement.

The training curves further illustrate this effect. VADER drives reward logits higher in later stages, alongside more pronounced declines in HPSv3 and Q-Align scores. Although quality decreases under both methods, TARO mitigates this deterioration (\cref{fig:bounded-vs-raw-rm}). Additional qualitative comparisons on Anime generation and neutral prompts are provided in the supplementary material.

\paragraph{Group-relative Policy Optimization.}
\label{sec:exp_tagrpo}

To evaluate TARO within group-relative policy optimization, we compare Flow-GRPO+TARO with Flow-GRPO on Wan2.1-T2V-1.3B. Flow-GRPO+TARO improves both target Mean and P20, indicating stronger and more consistent target expression (\cref{tab:tabo-ta-grpo-target}). It also improves the VBench mean, mainly through increased dynamic degree, while maintaining a comparable VideoScore2 mean. These results show that integrating TARO into Flow-GRPO enhances the target attribute without substantially compromising overall video quality. Additional results on neutral prompts are provided
in the supplementary material. The qualitative comparisons in \cref{fig:video-vis-GRPO-compare} further demonstrate the visual improvement: Flow-GRPO+TARO produces more consistent 2D anime appearance while retaining recognizable scene composition.

\section{Conclusion}
\label{sec:conclusion}

We propose GenTA, a reward-guided post-training framework for improving fine-grained visual-attribute instruction following in pretrained visual generators. GDRM constructs generator-relevant, attribute-discriminative reward supervision by combining calibrated generator supports with content-matched real comparisons. Building on GDRM, TARO introduces a smooth, upper-bounded reward objective calibrated using real target-positive examples, enabling controlled optimization through both direct reward-gradient and group-relative policy-based post-training. Advantage-level adaptation preserves TARO's target-dependent reward attenuation during group-relative optimization, while condition-dependent reference preservation helps retain pretrained generation behavior. Experiments on image and video generation demonstrate improved target-attribute expression while maintaining competitive generation quality across both modalities. Together, GDRM establishes generator-relevant, attribute-discriminative reward supervision, while TARO controls the strength with which this supervision drives optimization toward a data-grounded target level.

\section{LLM Usage Statement}
ChatGPT was used to assist with editing author-written text, related work and citations, code and implementation, and author-created figures and diagrams. No LLMs were used for experiments, data creation or annotation, or research ideation and conclusions. The manuscript was not submitted for reviewer-style feedback. The authors verified all content and take full responsibility for the submitted work.

{\small
\bibliographystyle{ieeenat_fullname}
\bibliography{references}
}

\clearpage
\maketitlesupplementary
\appendix
\numberwithin{equation}{section}
\numberwithin{figure}{section}
\numberwithin{table}{section}

\section{Generator-guided Discriminative Reward Modeling}
\label{app:gdrm}

\subsection{Generator-Support Calibration and Supervision Construction}
\label{app:gdrm_data}

\paragraph{Shared latent visual descriptors.}
We represent real samples and generated anchors with modality-specific descriptors computed from their VAE latents and final denoised latents, respectively. These descriptors are used both to compute generator-support distances and to provide the visual-similarity branch of content matching. For an image latent $\mathbf z\in\mathbb R^{16\times64\times64}$, we concatenate channel-wise spatial means and standard deviations with an $8\times8$ adaptive average-pooled feature map:
\begin{equation}
    \begin{aligned}
    \mathbf d_{\mathrm{img}}(\mathbf z)
    =\bigl[&\boldsymbol\mu_{H,W}(\mathbf z);
    \boldsymbol\sigma_{H,W}(\mathbf z);\\
    &\operatorname{vec}\!\left(\operatorname{AAP}_{8\times8}(\mathbf z)\right)\bigr]
    \in\mathbb R^{1056}.
    \end{aligned}
\end{equation}
For a Wan2.1 video latent $\mathbf z\in\mathbb R^{16\times21\times60\times104}$, the descriptor concatenates channel-wise spatiotemporal means and standard deviations, an adaptively pooled $16\times4\times8\times8$ feature map, and the corresponding statistics of consecutive-frame differences:
\begin{equation}
    \begin{aligned}
    \mathbf d_{\mathrm{vid}}(\mathbf z)=\bigl[&\boldsymbol\mu_{T,H,W}(\mathbf z);
    \boldsymbol\sigma_{T,H,W}(\mathbf z);\\
    &\operatorname{vec}\!\left(\operatorname{AAP}_{4\times8\times8}(\mathbf z)\right);\\
    &\boldsymbol\mu_{T,H,W}(\Delta_t\mathbf z);
    \boldsymbol\sigma_{T,H,W}(\Delta_t\mathbf z)\bigr]
    \in\mathbb R^{4160},
    \end{aligned}
\end{equation}
where $\Delta_t\mathbf z$ denotes the difference between consecutive latent frames. We use $\mathbf d(x)$ to denote the descriptor for the corresponding modality and apply $\ell_2$ normalization throughout GDRM:
\begin{equation}
    \mathbf e_{\mathrm{vis}}(x)
    =\frac{\mathbf d(x)}{\|\mathbf d(x)\|_2}.
\end{equation}

\paragraph{Support-distance implementation.}
We compute support distances in the normalized descriptor space. For $s\in\{B,T\}$, let $\mathcal A_s$ denote the corresponding Base or Target anchor set. The distance from a real sample $x$ to support $s$ is
\begin{equation}
    \begin{aligned}
    d_s(x)&=\frac{1}{k}\sum_{u\in\operatorname{kNN}(x;\mathcal A_s)}
    \ell_{\mathrm{vis}}(x,u),\\
    \ell_{\mathrm{vis}}(x,u)&=1-\mathbf e_{\mathrm{vis}}(x)^\top
    \mathbf e_{\mathrm{vis}}(u).
    \end{aligned}
    \label{eq:app_support_distance}
\end{equation}
We set $k=10$ and independently calibrate the Base and Target support radii. For each anchor $u_i$, its leave-one-out radius excludes the anchor itself:
\begin{equation}
    r_s(u_i)=\frac{1}{k}\sum_{u\in\operatorname{kNN}(u_i;\mathcal A_s\setminus\{u_i\})}
    \ell_{\mathrm{vis}}(u_i,u).
    \label{eq:app_support_loo}
\end{equation}
The support threshold is the 95th percentile of these radii,
\begin{equation}
    \delta_s=Q_{0.95}\left(\{r_s(u_i):u_i\in\mathcal A_s\}\right).
\end{equation}
\begin{equation}
    \Gamma_s(x)=\mathbb{I}\left[d_s(x)\leq\delta_s\right].
    \label{eq:app_support_gate}
\end{equation}
We further define the normalized support distances and their relative margin as
\begin{equation}
    \rho_s(x)=\frac{d_s(x)}{\delta_s},\qquad
    m(x)=\rho_B(x)-\rho_T(x).
    \label{eq:app_support_margin}
\end{equation}
Membership is encoded by $\Gamma_s(x)$ in \cref{eq:app_support_gate}. The relative margin in \cref{eq:app_support_margin} is non-positive when a sample is at least as close to Base as to Target after radius calibration. Support membership organizes real examples by empirical generator relevance without changing their original attribute labels.

\paragraph{Generator-aligned supervision pools.}
Let $\mathcal{P}_a$ and $\mathcal{N}_a$ denote the real positive and negative sets. We retain the original binary labels. Support geometry is used only to assign samples to supervision pools and to identify ambiguous gray negatives for exclusion; it never relabels a sample. The matched-positive pool contains all positives supported by the Target anchors,
\begin{equation}
    \mathcal P_{\mathrm{match}}
    =
    \left\{
    x\in\mathcal{P}_a:
    \Gamma_T(x)=1
    \right\}.
\end{equation}
These matched positives can be described as target-preferred, base-preferred boundary, or target-only according to whether they are in both supports with $m(x)>0$, in both with $m(x)\leq0$, or in Target support alone, respectively. This descriptive subdivision does not remove or relabel any positive. Remaining positives form the coverage-positive pool $\mathcal P_{\mathrm{cov}}=\mathcal{P}_a\setminus\mathcal P_{\mathrm{match}}$.

Negative construction is asymmetric and follows a fixed order. We first remove gray negatives supported by Target but not by Base,
\begin{equation}
    \mathcal N_{\mathrm{gray}}
    =
    \left\{
    x\in\mathcal{N}_a:
    \Gamma_T(x)=1,\;\Gamma_B(x)=0
    \right\}.
\end{equation}
We remove $\mathcal N_{\mathrm{gray}}$ to form $\mathcal N_{\mathrm{elig}}=\mathcal N_a\setminus\mathcal N_{\mathrm{gray}}$. We retrieve $\mathcal N_{\mathrm{hard}}\subseteq\mathcal N_{\mathrm{elig}}$ using \cref{app:content_matching}, then form $\mathcal N_{\mathrm{rem}}=\mathcal N_{\mathrm{elig}}\setminus\mathcal N_{\mathrm{hard}}$. The matched-negative pool is
\begin{equation}
    \mathcal N_{\mathrm{match}}
    =
    \left\{\begin{aligned}
    x\in\mathcal N_{\mathrm{rem}}:
    &\Gamma_B(x)=1,\\
    &[\Gamma_T(x)=0\ \lor\ m(x)\leq0]
    \end{aligned}\right\},
\end{equation}
and the coverage-negative pool is $\mathcal N_{\mathrm{cov}}=\mathcal N_{\mathrm{rem}}\setminus\mathcal N_{\mathrm{match}}$. These five pools are disjoint: $\mathcal P_{\mathrm{match}}$ and $\mathcal P_{\mathrm{cov}}$ partition all labeled positives, while the three negative pools partition $\mathcal N_{\mathrm{elig}}$ after gray-negative exclusion. No real positive is removed by the support gates. Pool membership changes sampling exposure, not binary label semantics.

\paragraph{Latent reward learning.}
For each real training sample $x$, the reward model receives its VAE latent $\mathbf z_{\mathrm{VAE}}(x)$ and is trained with binary cross-entropy:
\begin{equation}
    \begin{aligned}
    \mathcal{L}_{\mathrm{RM}}
    &=-y\log\sigma\!\left(R_\phi^a(\mathbf z_{\mathrm{VAE}}(x))\right)\\
    &\quad -(1-y)\log\!\left(1-\sigma\!\left(R_\phi^a(\mathbf z_{\mathrm{VAE}}(x))\right)\right).
    \end{aligned}
\end{equation}
Here $R_\phi^a(\mathbf z_{\mathrm{VAE}}(x))$ is the raw reward logit, and the sigmoid converts it to a probability only for the binary cross-entropy objective. During generator post-training, the same reward model receives the generated sample's final denoised latent instead of a real-sample VAE encoding. The reward model is frozen during target calibration and generator post-training.

\subsection{Multimodal Content-Matched Real Attribute Boundary}
\label{app:content_matching}

\paragraph{Target-masked structured-tag similarity.}
Each sample is represented by structured tags organized into semantic fields. Let $\mathcal{F}$ denote the set of fields. A field represents a semantic factor, such as \texttt{subject}, \texttt{scene}, \texttt{style}, or \texttt{motion}, and its tags describe the values observed for that factor. For example, \texttt{scene} may contain tags such as \texttt{street} and \texttt{night}, while \texttt{style} may contain \texttt{oil painting}. Let $f_a$ be the field containing attribute $a$. We mask the entire target field during semantic matching,
\begin{equation}
    \mathcal{F}_{-a}=\mathcal{F}\setminus\{f_a\},
\end{equation}
so the target attribute cannot directly determine the retrieved neighbors. A tag $t$ retains its category, subcategory, and tag name when encoded:
\begin{equation}
\begin{aligned}
    \mathbf{e}_{\mathrm{tag}}(t)
    &=\operatorname{Normalize}\bigl(\operatorname{Enc}_{\mathrm{tag}}(
    \operatorname{category}(t),\\
    &\qquad\operatorname{subcategory}(t),
    \operatorname{name}(t))\bigr),\\
    s(t,t')
    &=\mathbf{e}_{\mathrm{tag}}(t)^\top
    \mathbf{e}_{\mathrm{tag}}(t').
\end{aligned}
\end{equation}
For a field $f$ shared by two samples, let $A_f$ and $B_f$ be their tag sets. We use a bidirectional best-match score,
\begin{equation}
    S_f(A,B)
    =
    \frac{1}{2}
    \left[\begin{aligned}
    \frac{1}{|A_f|}
    &\sum_{t\in A_f}\max_{t'\in B_f}s(t,t')\\
    {}+\frac{1}{|B_f|}
    &\sum_{t'\in B_f}\max_{t\in A_f}s(t,t')
    \end{aligned}\right].
\end{equation}
The first term averages the best match for each tag in $A_f$ against $B_f$, and the second performs the reverse matching. Averaging both directions prevents a sample with a richer tag set from being considered fully matched based on only a few tags in the other sample. Let $\mathcal{F}(A,B)$ be the non-empty fields shared by $A$ and $B$ after masking $f_a$, and let $\mathcal{F}_A$ and $\mathcal{F}_B$ be their corresponding non-empty field sets. The structured similarity and field coverage are
\begin{equation}
    \begin{aligned}
    S_{\mathrm{tag}}(A,B)
    &=\frac{1}{|\mathcal{F}(A,B)|}
    \sum_{f\in\mathcal{F}(A,B)}S_f(A,B),\\
    \operatorname{Cov}(A,B)
    &=\frac{|\mathcal{F}(A,B)|}
    {|\mathcal{F}_A\cup\mathcal{F}_B|}.
    \end{aligned}
\end{equation}
We retain a pair for fine-grained scoring only if it shares at least three non-empty non-target fields and has field coverage of at least $0.30$. This requirement prevents a high similarity score from being supported by only a small subset of the available non-target semantics.

\paragraph{Distribution-calibrated fusion.}
The visual similarity is the cosine similarity of the shared normalized descriptors, $S_{\mathrm{img}}(p,n)=\mathbf e_{\mathrm{vis}}(p)^\top\mathbf e_{\mathrm{vis}}(n)$. For each target-attribute task, we form candidate edges from real positives $p$ to eligible real negatives $n$. Because each task uses a single data modality, we estimate one empirical cumulative distribution function (CDF) for each similarity branch over these positive-to-negative pairs. The two similarities are mapped to their empirical quantiles:
\begin{equation}
    \begin{aligned}
    Q_{\mathrm{tag}}(p,n)
    &=\widehat F_{\mathrm{tag}}\bigl(S_{\mathrm{tag}}(p,n)\bigr),\\
    Q_{\mathrm{img}}(p,n)
    &=\widehat F_{\mathrm{img}}\bigl(S_{\mathrm{img}}(p,n)\bigr).
    \end{aligned}
\end{equation}
We fuse the calibrated similarities using a weighted geometric mean,
\begin{equation}
    S_{\mathrm{fused}}(p,n)
    =Q_{\mathrm{tag}}(p,n)^w
    Q_{\mathrm{img}}(p,n)^{1-w},
    \qquad w=0.5.
\end{equation}
Thus, structured-tag and visual similarity contribute equally. We estimate the empirical CDFs independently for each target-attribute task, making the two branches comparable without requiring their raw scores to share a common numerical scale.

\paragraph{Coarse-to-fine retrieval.}
For each labeled positive $p\in\mathcal P_a$, we retrieve candidates exclusively from the eligible negative set $\mathcal N_{\mathrm{elig}}$. We obtain the coarse semantic representation $h(x)$ by averaging the representations of the non-target fields and applying $\ell_2$ normalization. The coarse candidate set is
\begin{equation}
    \mathcal C_{\mathrm{coarse}}(p)
    =\operatorname{TopK}_{256}
    \left(\mathcal N_{\mathrm{elig}};
    \,h(p)^\top h(n)\right),
\end{equation}
We next apply the field-overlap gate, compute the tag and visual similarities, calibrate both scores, and rank the remaining candidates by $S_{\mathrm{fused}}$. For each positive, we retain the 10 highest-ranked candidates before thresholding; consequently, fewer than 10 negatives may remain. The retrieval stages and essential settings are summarized in \cref{tab:gdrm_retrieval_pipeline}.

\begin{table*}[tbp]
    \centering
    \small
    \caption{Coarse-to-fine retrieval of content-matched hard negatives in GDRM. The procedure is applied independently to each target-attribute task and its data modality.}
    \label{tab:gdrm_retrieval_pipeline}
    \setlength{\tabcolsep}{5pt}
    \renewcommand{\arraystretch}{1.12}
    \begin{tabular}{p{0.15\textwidth}p{0.31\textwidth}p{0.43\textwidth}}
        \toprule
        Stage & Operation & Criterion / setting \\
        \midrule
        1. Coarse retrieval & For each positive, retrieve negative candidates using cosine similarity between non-target semantic descriptors. & Top $K_{\mathrm{coarse}}=256$ candidates from $\mathcal N_{\mathrm{elig}}$. \\
        2. Field-overlap gate & Measure the number of shared non-empty non-target fields and their coverage. & Require at least 3 shared fields and $\operatorname{Cov}\geq0.30$. \\
        3. Fine scoring & Compute bidirectional structured-tag similarity and normalized latent visual similarity. & Obtain $S_{\mathrm{tag}}$ and $S_{\mathrm{img}}$ for each remaining positive--negative pair. \\
        4. Calibration and fusion & Map each similarity to its task-specific empirical quantile and combine the two quantiles. & $w=0.5$ in $S_{\mathrm{fused}}=Q_{\mathrm{tag}}^{w}Q_{\mathrm{img}}^{1-w}$. \\
        5. Per-positive ranking & Rank candidates for each positive by fused similarity. & Retain the top $K_{\mathrm{hard}}=10$ candidates before threshold filtering. \\
        6. Threshold filtering & Filter the top-ranked candidates and aggregate the retained negatives. & $Q_{\mathrm{tag}}\geq0.90$, $Q_{\mathrm{img}}\geq0.90$, and $S_{\mathrm{fused}}\geq0.90$. \\
        \bottomrule
    \end{tabular}
\end{table*}

\begin{figure*}[tbp]
    \centering
    \includegraphics[width=0.67\textwidth,height=0.76\textheight,keepaspectratio]{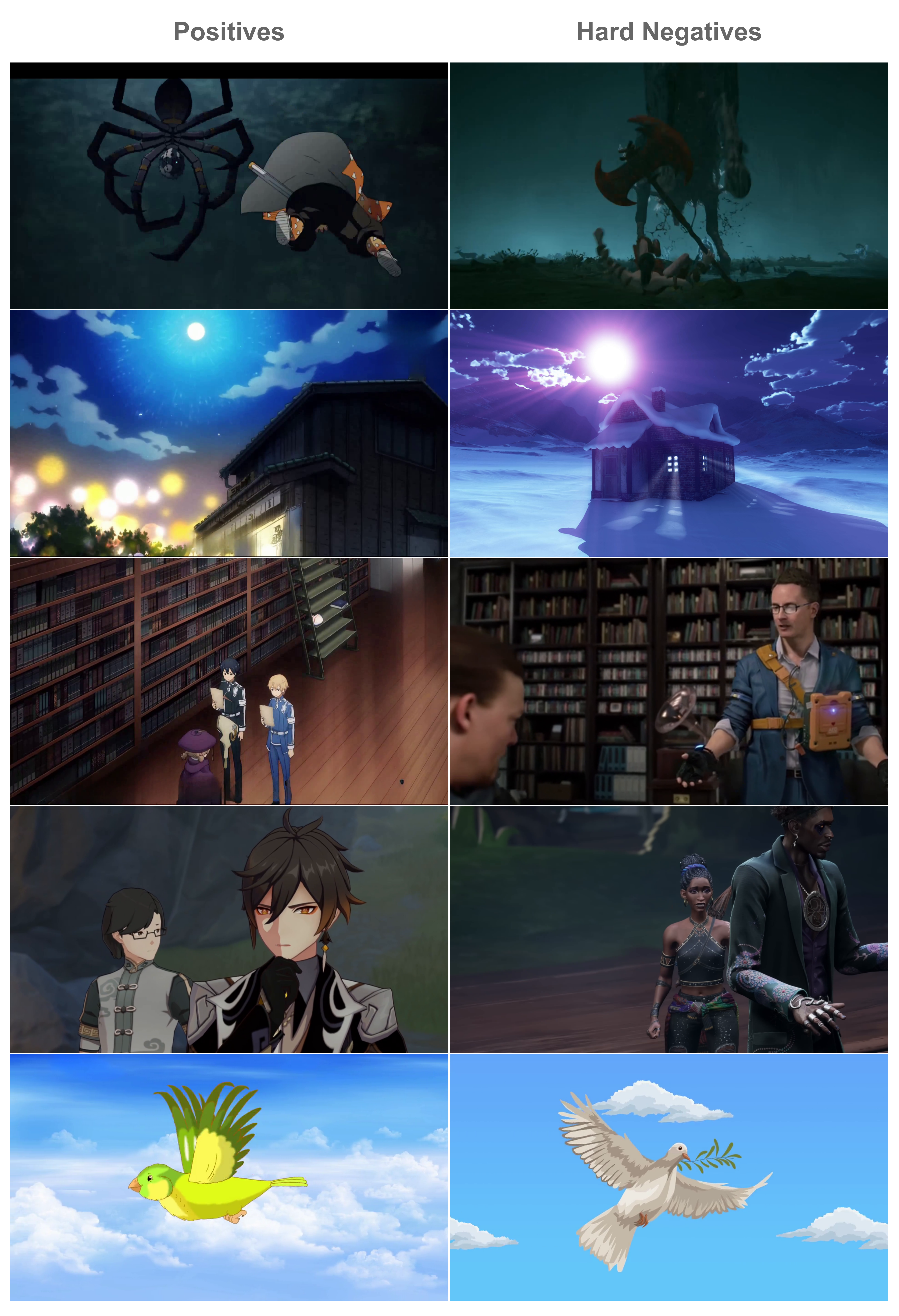}
    \caption{\textbf{Qualitative examples of content-matched Anime hard negatives.} Each row shows a real positive (left) and its retrieved hard negative (right). The pairs preserve related non-target content while exhibiting different Anime-style expression; exact scene correspondence is not required.}
    \label{fig:anime_content_matched_hard_negatives}
\end{figure*}

\paragraph{Qualitative content-matching examples.}
As shown in \cref{fig:anime_content_matched_hard_negatives}, the retrieved hard negatives preserve broad semantic content similar to their corresponding Anime-positive samples, including underwater scenes, nighttime buildings, libraries, human subjects, and flying birds. Although the pairs are not perfectly content-aligned, they share meaningful non-target visual and semantic characteristics while differing in the target Anime style. These examples illustrate how content matching helps reduce content confounding and construct more attribute-discriminative reward supervision.

\section{Target-Anchored Reward Optimization}
\label{app:taro}

\subsection{Reward Calibration and Direct Reward-Gradient Optimization}
\label{app:taro_calibration}

\paragraph{Calibration from real positive training examples.}
After freezing $R_\phi^a$, we calibrate TARO using real positive examples from the reward-model training split. Let $\mathcal R_a^+$ denote the corresponding set of reward logits:
\begin{equation}
    \begin{aligned}
    \mathcal R_a^+&=\left\{R_\phi^a(\mathbf z_{\mathrm{VAE}}(x)):
    x\in\mathcal D^+_{a,\mathrm{train}}\right\},\\
    T_a&=Q_{0.75}(\mathcal R_a^+),\\
    \eta_a&=\tfrac12\operatorname{IQR}(\mathcal R_a^+)+10^{-6}.
    \end{aligned}
\end{equation}
We fix the calibrated target $T_a$ and transition width $\eta_a$ throughout post-training. The target $T_a$ serves as a soft saturation reference in reward space, not as a hard stopping threshold.

TARO transforms the raw reward logit $R$ using the smooth, upper-bounded function
\begin{equation}
    \begin{aligned}
    B_a(R;T_a,\eta_a)
    &=-\eta_a\operatorname{softplus}\!\left(\frac{T_a-R}{\eta_a}\right),\\
    \frac{\partial B_a}{\partial R}
    &=\sigma\!\left(\frac{T_a-R}{\eta_a}\right).
    \end{aligned}
\end{equation}
Well below $T_a$, the transformed reward approaches $R-T_a$ and its slope remains close to one. As $R$ approaches and exceeds the target, the slope decays smoothly toward zero and the transformed reward approaches its upper bound of zero. TARO therefore preserves a strong optimization signal when the target attribute is under-expressed while progressively reducing the incentive for further reward increases near and beyond the calibrated target.

\paragraph{VADER reward-gradient realization.}
For differentiable optimization with VADER, TARO enters the generator update through the final denoised latent. Let $\mathbf z^{\mathrm{final}}_\theta$ denote this latent for a generated sample. The resulting gradient is
\begin{equation}
    \begin{aligned}
    \nabla_\theta B_a
    ={}&\frac{\partial B_a}{\partial R_\phi^a}
    \frac{\partial R_\phi^a(\mathbf z^{\mathrm{final}}_\theta)}
    {\partial\mathbf z^{\mathrm{final}}_\theta}\\
    &\cdot\frac{\partial\mathbf z^{\mathrm{final}}_\theta}{\partial\theta}.
    \end{aligned}
\end{equation}
The first factor attenuates the reward gradient as the generated reward approaches the calibrated target; the remaining gradient path is unchanged from the underlying differentiable optimization method.

\begin{figure*}[t]
    \centering
    \includegraphics[width=0.78\textwidth]{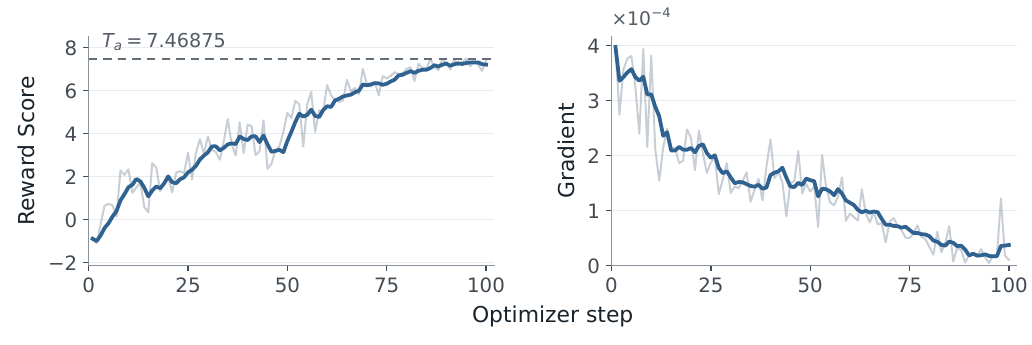}
    \caption{\textbf{Reward and gradient dynamics during Anime video post-training.} With Wan2.1-T2V-14B and VADER+TARO, the generated raw latent reward score (left) approaches the calibrated target $T_a=7.46875$ (dashed line) as the scaled reward-gradient norm in prediction space (right) decreases over 100 optimizer steps.}
    \label{fig:anime_taro_reward_gradient}
\end{figure*}

\paragraph{Observed reward and gradient dynamics.}
As shown in \Cref{fig:anime_taro_reward_gradient}, during Anime post-training with VADER+TARO, the raw latent reward approaches the real-positive-calibrated target $T_a=7.46875$, while the scaled reward-gradient norm in prediction space decreases. This trend is consistent with TARO attenuating reward-driven updates near the target.

\subsection{TARO for Group-Relative Policy Optimization}
\label{app:taro_flow_grpo}

In Flow-GRPO+TARO, we transform each sample reward as $\widetilde R_i=B_a(R_i;T_a,\eta_a)$. We center the transformed rewards within each prompt group $G$ and normalize them using a shared scale $v_k$ across groups in the same update interval:
\begin{equation}
    A_i^{\mathrm{TARO}}
    =\frac{\widetilde R_i-\frac{1}{|G|}\sum_{j\in G}\widetilde R_j}
    {v_k+\epsilon}.
\end{equation}
Using a shared scale instead of each group's standard deviation preserves the attenuation introduced by TARO as group rewards approach the calibrated target. We initialize $v_0$ as the median within-group standard deviation across four calibration groups, each containing eight samples. Every 25 optimizer updates, we refresh the scale using the median within-group standard deviation $\sigma_g$ across groups $\mathcal G_k$ from the preceding interval. The update uses an exponential moving average with coefficient $\alpha=0.2$ and a lower bound of $0.5v_0$, where $\sigma_g$ is the standard deviation of transformed rewards within group $g$:
\begin{equation}
    v_{k+1}=\max\!\left(0.5v_0,\;0.8v_k+0.2\,\operatorname{median}_{g\in\mathcal G_k}\sigma_g\right).
\end{equation}
The same initialization and update configuration is used across tasks.

\paragraph{Timestep balancing.}
When enabled, timestep balancing uses fixed weights to compensate for differences in typical gradient magnitudes across selected flow transitions. Let $g_t$ denote the median gradient magnitude at transition $t$, and define $g_{\mathrm{ref}}=\operatorname{median}_t g_t$. The unnormalized weight is
\begin{equation}
    \begin{aligned}
    r_t&=\sqrt{\frac{g_{\mathrm{ref}}}{g_t+10^{-12}}},\\
    w_t&=\operatorname{clip}(r_t,0.25,4.0).
    \end{aligned}
\end{equation}
For the selected set $\mathcal S$ with $K=8$, we normalize the weights as $\bar w_t=w_t/(K^{-1}\sum_{j\in\mathcal S}w_j)$ and apply each weight once to its corresponding policy-gradient contribution. Clipping is performed before normalization, so the normalized weights are not necessarily confined to $[0.25,4.0]$. This correction balances transition-level contributions without changing the video-level reward or the TARO transformation.

\section{Additional Experimental Setup}
\label{app:additional_setup}

\subsection{Tasks, Data, and Attribute Triggers}

We organize the setup into three parts: real-data construction and pool sampling, latent reward-model training and validation, and independent evaluation in RGB pixel space.

We study three image attributes---Vintage Faded, Oil Painting, and Muscular---using Z-Image. We also study 2D Japanese Anime video generation using Wan2.1-T2V. The video experiments use Wan2.1-T2V-14B with VADER and Wan2.1-T2V-1.3B with Flow-GRPO. We specify the target conditions using the following tag-trigger strings:

\begin{table*}[tbp]
    \centering
    \small
    \caption{Tag-trigger strings used to specify the target conditions. The text is reproduced verbatim.}
    \label{tab:attribute_triggers}
    \setlength{\tabcolsep}{6pt}
    \renewcommand{\arraystretch}{1.12}
    \begin{tabular*}{\linewidth}{@{\extracolsep{\fill}}p{0.22\linewidth}p{0.70\linewidth}@{}}
        \toprule
        Target attribute & Trigger string \\
        \midrule
        Vintage Faded & \texttt{This image has vintage faded style.} \\
        Oil Painting & \texttt{This image has a oil painting style.} \\
        Muscular & \texttt{The person or people in the image have a visibly muscular build.} \\
        2D Japanese Anime (video) & \texttt{2D Japanese anime animation.} \\
        \bottomrule
    \end{tabular*}
\end{table*}

For image reward modeling, we construct approximately 60K image--text pairs from Re-LAION-Caption19M through quality filtering and seeded sampling, including roughly 1K target-positive examples. The video corpus contains approximately 5.5M videos with structured annotations. For Anime reward modeling, we combine approximately 10K target-positive videos with negatives sampled from the full negative pool, yielding about 100K training examples. These real-data collections are used to train the reward models and are separate from the prompts used for generator post-training. Post-training prompts are drawn from DanceGRPO. We evaluate each image attribute on 300 prompts and Anime on 200 prompts. Muscular prompts are restricted to people and settings suitable for assessing muscular definition.

\subsection{GDRM Pool Sampling}

We sample from the five GDRM pools using the modality-specific proportions reported in \cref{tab:pool_ratios}.

\begin{table}[!tbp]
    \centering
    \small
    \caption{Sampling proportions for the five GDRM training pools.}
    \label{tab:pool_ratios}
    \begin{tabular}{lcc}
        \toprule
        Pool & Image & Video \\
        \midrule
        Matched positive & 0.380 & 0.400 \\
        Coverage positive & 0.120 & 0.100 \\
        Content-matched hard negative & 0.075 & 0.160 \\
        Matched negative & 0.275 & 0.240 \\
        Coverage negative & 0.150 & 0.100 \\
        \bottomrule
    \end{tabular}
\end{table}

\subsection{Latent Reward Models and Anchor Sets}
\label{app:latent_rm}

The image reward model is a six-layer latent ViT with patch size 4, hidden dimension 384, and six attention heads, followed by a linear reward head. The video reward model processes Wan latents of shape $[B,16,21,60,104]$ using local spatial tokenization and temporal attention, and predicts a scalar logit from the aggregated video representation. Both models are trained on real-sample VAE latents and receive final denoised latents during generator post-training.

For GDRM support construction, we generate matched Base/Target anchor pairs while holding the content prompt, seed, checkpoint, and sampling settings fixed; only the target-attribute condition changes. We use 5K pairs for Vintage Faded, Oil Painting, and Anime, and 2.5K pairs for Muscular. The anchors are stored as final denoised latents and define the generator-native supports.

For reward-model checkpoint selection, we use separate held-out Base/Target pairs: 2K pairs for each image attribute and 250 pairs for Anime. These validation pairs are disjoint from both the support anchors and the real positives used to calibrate TARO.

\subsection{Latent Reward-Model Validation}
\label{app:latent_rm_validation}

We select latent reward-model checkpoints using the held-out generated pairs in \cref{tab:latent_rm_validation}. Preference rate is the fraction of pairs in which the Target sample receives a higher reward than its matched Base sample. Target--Base AUROC measures the separation between the two score distributions. We prioritize preference rate during checkpoint selection and use AUROC to break near ties; the raw logit margin is reported only as a diagnostic.

\par\medskip
\noindent\begin{minipage}{\linewidth}
    \centering
    \small
    \captionsetup{hypcap=false}
    \captionof{table}{Held-out generated-pair validation for the latent reward models.}
    \label{tab:latent_rm_validation}
    \setlength{\tabcolsep}{3pt}
    \begin{tabular}{@{}lrrr@{}}
        \toprule
        Modality / tag & \shortstack[r]{Pair\\count} & \shortstack[r]{Preference\\rate}
                       & \shortstack[r]{Target--Base\\AUROC} \\
        \midrule
        Image / Vintage Faded & 2,000 & 0.9265 & 0.823734 \\
        Image / Oil Painting & 2,000 & 0.9395 & 0.841797 \\
        Image / Muscular & 2,000 & 0.7455 & 0.686098 \\
        Video / Anime & 250 & 0.9200 & 0.871304 \\
        \bottomrule
    \end{tabular}
\end{minipage}

\subsection{RGB Evaluators and Evaluation Protocol}
\label{app:evaluators}

We train separate evaluators in RGB pixel space to measure target-attribute expression and freeze them during generator post-training. These RGB-pixel-space evaluators are distinct from the latent reward models. The image evaluator uses a frozen SigLIP vision encoder with a linear classification head. The video evaluator aggregates frozen SigLIP features from eight frames sampled from each five-second clip using a lightweight temporal Transformer and a learned video token.

The image evaluator uses a disjoint 80/20 train/validation split on real data (\cref{tab:image_evaluator_split}); held-out generated Base/Target pairs are reserved for evaluation in the generated domain. The video evaluator is split by video identity into disjoint train, validation, and test sets in an 80/10/10 ratio (\cref{tab:video_evaluator_split}).

\noindent\begin{minipage}{\linewidth}
Clips labeled with other animation styles are hard negatives; clips without any animation-style label are negatives. Original sampling categories are retained.
\end{minipage}

\begin{table*}[!tbp]
    \centering
    \small
    \caption{Image RGB-evaluator train/validation composition.}
    \label{tab:image_evaluator_split}
    \setlength{\tabcolsep}{3pt}
    \begin{tabular*}{\linewidth}{@{\extracolsep{\fill}}lrrrrrr@{}}
        \toprule
        Tag & \multicolumn{3}{c}{Train} & \multicolumn{3}{c}{Validation} \\
        \cmidrule(lr){2-4} \cmidrule(lr){5-7}
        & Pos. & \shortstack[r]{Content-matched\\hard neg.} & Coverage neg.
        & Pos. & \shortstack[r]{Content-matched\\hard neg.} & Coverage neg. \\
        \midrule
        Vintage Faded & 1,108 & 10,114 & 36,613 & 278 & 2,476 & 9,206 \\
        Oil Painting & 548 & 6,010 & 41,220 & 134 & 1,466 & 10,297 \\
        Muscular & 621 & 6,204 & 40,575 & 185 & 1,529 & 10,211 \\
        \bottomrule
    \end{tabular*}
\end{table*}

\begin{table}[!tbp]
    \centering
    \small
    \caption{Video RGB-evaluator split composition.}
    \label{tab:video_evaluator_split}
    \begin{tabular}{lrrr}
        \toprule
        Split & Total & Positive & Negative \\
        \midrule
        Train & 73,069 & 7,306 & 65,763 \\
        Validation & 9,133 & 913 & 8,220 \\
        Test & 9,134 & 913 & 8,221 \\
        \midrule
        Total & 91,336 & 9,132 & 82,204 \\
        \bottomrule
    \end{tabular}
\end{table}

We select generator checkpoints using the mean RGB-evaluator logit on held-out DanceGRPO prompts. This prompt-level validation set is distinct from both the real-data evaluator split and the generated pairs used for latent reward-model checkpoint selection. Target strength is reported as the mean and 20th percentile (P20) of RGB scores mapped to percentiles using Target-validation anchors. Image quality is evaluated with HPSv2, HPSv3, PickScore, VQAScore, and Q-Align quality and aesthetics. For video, the primary VBench score averages subject consistency, dynamic degree, and aesthetic quality, while the VideoScore2 score averages visual quality, text alignment, and physical consistency. Additional six-component VBench and VideoReward diagnostics are reported in \cref{app:video_post_target_full}.

\section{Additional Experimental Results}
\label{app:additional_results}

\subsection{Complete GDRM Component Ablations}
\label{app:gdrm_component_ablations}

Our primary comparison includes Direct Real, Generated-only, and GDRM. The complete image and video results in \cref{tab:image-post-target-complete,tab:video-post-target-complete} further isolate the effects of content matching and generator-support calibration through the Content-Matched-only and Support-only variants.

\begin{table*}[tbp]
\centering
\caption{\textbf{Post-target image-generation results under different reward-supervision constructions.} Target strength is measured by RGB-score percentiles calibrated on the Target-validation anchors. Base is shown only as a reference; \textbf{bold} and underlining indicate the best and second-best post-trained results, respectively, for each metric.}
\label{tab:image-post-target-complete}
\small
\setlength{\tabcolsep}{2.6pt}
\renewcommand{\arraystretch}{1.06}
\begin{tabular*}{\linewidth}{@{\extracolsep{\fill}}lrrrrrrrr@{}}
\toprule
Method & Mean $\uparrow$ & P20 $\uparrow$ & HPSv3 $\uparrow$ & Q-Align Q. $\uparrow$ & Q-Align Aes. $\uparrow$ & HPSv2 $\uparrow$ & PickScore $\uparrow$ & VQA $\uparrow$ \\
\midrule
\multicolumn{9}{c}{\textbf{\textit{Muscular}}} \\
\addlinespace[1pt]
\rowcolor{gray!10}
Base (ref.) & 51.838 & 24.835 & 9.469 & 4.826 & 3.881 & 0.307 & 20.582 & 0.706 \\
Direct Real & 57.341 & 29.196 & 9.340 & 4.753 & 3.778 & 0.308 & 20.554 & 0.684 \\
Generated-only & 55.034 & 31.043 & 8.983 & 4.568 & 3.534 & 0.305 & 20.507 & 0.680 \\
Content-Matched-only & 57.445 & 29.880 & 9.427 & \underline{4.812} & \textbf{3.915} & \underline{0.310} & 20.564 & \underline{0.700} \\
Support-only & \underline{59.173} & \underline{31.807} & \textbf{9.551} & 4.703 & 3.754 & \underline{0.310} & \underline{20.602} & 0.697 \\
GDRM & \textbf{62.862} & \textbf{36.896} & \underline{9.549} & \textbf{4.818} & \underline{3.888} & \textbf{0.312} & \textbf{20.673} & \textbf{0.720} \\

\addlinespace[3pt]
\midrule
\addlinespace[1pt]
\multicolumn{9}{c}{\textbf{\textit{Oil Painting}}} \\
\addlinespace[1pt]
\rowcolor{gray!10}
Base (ref.) & 51.282 & 26.436 & 10.818 & 4.416 & 3.798 & 0.299 & 20.235 & 0.730 \\
Direct Real & \underline{74.455} & 55.177 & \underline{9.787} & \underline{3.912} & 3.636 & 0.282 & \underline{19.992} & \underline{0.687} \\
Generated-only & 67.135 & 45.234 & 9.226 & 3.728 & 3.474 & 0.277 & 19.969 & 0.684 \\
Content-Matched-only & 73.529 & \underline{55.847} & 9.418 & 3.887 & 3.609 & 0.274 & 19.938 & 0.677 \\
Support-only & 73.849 & 53.438 & 9.690 & 3.898 & \underline{3.640} & \underline{0.284} & 19.989 & 0.683 \\
GDRM & \textbf{76.102} & \textbf{58.436} & \textbf{10.146} & \textbf{4.043} & \textbf{3.696} & \textbf{0.291} & \textbf{20.074} & \textbf{0.697} \\

\addlinespace[3pt]
\midrule
\addlinespace[1pt]
\multicolumn{9}{c}{\textbf{\textit{Vintage Faded}}} \\
\addlinespace[1pt]
\rowcolor{gray!10}
Base (ref.) & 50.733 & 22.606 & 11.555 & 4.447 & 3.624 & 0.294 & 19.857 & 0.708 \\
Direct Real & 72.862 & 54.527 & 10.108 & 3.883 & 3.377 & 0.288 & 19.949 & \textbf{0.710} \\
Generated-only & 62.153 & 33.596 & \textbf{10.851} & \textbf{4.146} & 3.488 & 0.285 & 19.987 & 0.671 \\
Content-Matched-only & 67.835 & 43.120 & \underline{10.842} & \underline{4.095} & 3.479 & \underline{0.291} & \underline{19.991} & 0.694 \\
Support-only & \textbf{75.614} & \textbf{58.356} & 10.271 & 3.929 & \textbf{3.506} & 0.287 & \textbf{19.997} & 0.696 \\
GDRM & \underline{74.425} & \underline{54.624} & 10.805 & 4.056 & \underline{3.497} & \textbf{0.292} & 19.971 & \underline{0.709} \\
\bottomrule
\end{tabular*}
\end{table*}

\begin{table*}[t]
\centering
\caption{\textbf{Post-target enhancement on Anime video (Wan2.1-T2V-14B).} Mean and P20 denote RGB target-strength percentiles calibrated on Target-validation anchors. SC, DD, and Aes. denote VBench subject consistency, dynamic degree, and aesthetic quality; VQ, TA, and PC denote VideoScore2 visual quality, text alignment, and physical consistency. Each reported mean is the arithmetic mean of its three corresponding metrics. Base is shown only as a reference; \textbf{bold} and underlining indicate the best and second-best post-trained values, respectively, with ties determined at the displayed precision.}
\label{tab:video-post-target-complete}
\small
\setlength{\tabcolsep}{2.6pt}
\renewcommand{\arraystretch}{1.06}
\begin{tabular*}{\linewidth}{@{\extracolsep{\fill}}lrrrrrrrrrr@{}}
\toprule
 & \multicolumn{2}{c}{Attribute} & \multicolumn{4}{c}{VBench} & \multicolumn{4}{c}{VideoScore2} \\
\cmidrule(lr){2-3}\cmidrule(lr){4-7}\cmidrule(lr){8-11}
Method & Mean $\uparrow$ & P20 $\uparrow$ & SC $\uparrow$ & DD $\uparrow$ & Aes. $\uparrow$ & Mean $\uparrow$ & VQ $\uparrow$ & TA $\uparrow$ & PC $\uparrow$ & Mean $\uparrow$ \\
\midrule
\rowcolor{gray!10}
Base (ref.) & 59.263 & 43.430 & 0.9725 & 0.2117 & 0.6655 & 0.6166 & 3.9720 & 3.8826 & 3.7742 & 3.8763 \\
Direct Real & 68.579 & \textbf{55.417} & 0.9717 & 0.1350 & 0.6362 & 0.5810 & 3.9586 & \textbf{4.1870} & 3.7643 & \underline{3.9700} \\
Generated-only & 68.079 & 53.974 & 0.9684 & 0.1150 & 0.6288 & 0.5707 & \underline{3.9713} & 3.9709 & 3.7016 & 3.8813 \\
Content-Matched-only & 65.589 & 51.195 & \underline{0.9770} & \underline{0.1400} & \textbf{0.6444} & \underline{0.5871} & 3.9384 & 4.0143 & 3.7400 & 3.8976 \\
Support-only & \textbf{69.160} & \underline{54.704} & 0.9766 & 0.0900 & 0.6356 & 0.5674 & 3.9371 & 3.9973 & \underline{3.8032} & 3.9125 \\
GDRM & \underline{68.722} & 54.267 & \textbf{0.9772} & \textbf{0.1450} & \underline{0.6434} & \textbf{0.5885} & \textbf{4.0182} & \underline{4.1391} & \textbf{3.8467} & \textbf{4.0013} \\
\bottomrule
\end{tabular*}
\end{table*}

\subsection{Neutral-Prompt Preservation}
\label{app:taro_neutral}

We evaluate neutral prompts using the same GDRM checkpoints as in the primary experiments. Across the image attributes, GDRM maintains low target-attribute leakage while retaining competitive generation quality. For Anime, Support-only yields the lowest post-trained leakage, while GDRM remains close and achieves the highest post-trained subject consistency and text-alignment scores. The optimizer comparisons in \cref{tab:taro-optimizer-neutral} show that TARO reduces neutral-prompt leakage with both optimizers: from 0.0341 to 0.0101 for VADER and from 0.1425 to 0.0631 for Flow-GRPO.

\begin{table*}[tbp]
\centering
\caption{\textbf{Neutral-prompt preservation for image generation under different reward-supervision constructions.} Leakage is the increase in predicted target-tag probability relative to Base (lower is better). Base is shown only as a reference; \textbf{bold} and underlining indicate the best and second-best post-trained results, respectively, for each metric.}
\label{tab:image-post-neutral}
\small
\setlength{\tabcolsep}{3.0pt}
\renewcommand{\arraystretch}{1.06}
\begin{tabular*}{\linewidth}{@{\extracolsep{\fill}}lrrrrrrr@{}}
\toprule
Method & Leakage $\downarrow$ & HPSv3 $\uparrow$ & Q-Align Q. $\uparrow$ & Q-Align Aes. $\uparrow$ & HPSv2 $\uparrow$ & PickScore $\uparrow$ & VQA $\uparrow$ \\
\midrule
\multicolumn{8}{c}{\textbf{\textit{Muscular}}} \\
\addlinespace[1pt]
\rowcolor{gray!10}
Base (ref.) & 0.000 & 10.847 & 4.751 & 3.765 & 0.305 & 20.356 & 0.786 \\
Direct Real & \underline{0.011} & 10.678 & 4.696 & 3.667 & \underline{0.306} & 20.328 & 0.783 \\
Generated-only & \textbf{0.002} & 10.584 & 4.618 & 3.568 & 0.304 & 20.307 & \underline{0.788} \\
Content-Matched-only & 0.015 & 10.779 & \textbf{4.743} & \textbf{3.770} & \textbf{0.308} & 20.323 & 0.785 \\
Support-only & 0.020 & \textbf{10.823} & 4.653 & 3.628 & \underline{0.306} & \underline{20.354} & 0.786 \\
GDRM & \textbf{0.002} & \underline{10.811} & \underline{4.739} & \underline{3.731} & \textbf{0.308} & \textbf{20.376} & \textbf{0.790} \\

\addlinespace[3pt]
\midrule
\addlinespace[1pt]
\multicolumn{8}{c}{\textbf{\textit{Oil Painting}}} \\
\addlinespace[1pt]
\rowcolor{gray!10}
Base (ref.) & 0.000 & 10.385 & 4.676 & 3.647 & 0.295 & 20.177 & 0.708 \\
Direct Real & 0.043 & \underline{10.376} & 4.624 & \underline{3.628} & 0.294 & 20.170 & 0.714 \\
Generated-only & \textbf{0.026} & 10.325 & 4.602 & 3.595 & 0.293 & 20.159 & \textbf{0.722} \\
Content-Matched-only & 0.042 & 10.363 & \underline{4.626} & 3.626 & 0.294 & \underline{20.176} & 0.712 \\
Support-only & 0.042 & 10.372 & 4.613 & 3.615 & \underline{0.295} & 20.164 & 0.717 \\
GDRM & \underline{0.036} & \textbf{10.456} & \textbf{4.636} & \textbf{3.640} & \textbf{0.296} & \textbf{20.187} & \underline{0.721} \\

\addlinespace[3pt]
\midrule
\addlinespace[1pt]
\multicolumn{8}{c}{\textbf{\textit{Vintage Faded}}} \\
\addlinespace[1pt]
\rowcolor{gray!10}
Base (ref.) & 0.000 & 10.385 & 4.676 & 3.647 & 0.295 & 20.177 & 0.708 \\
Direct Real & 0.046 & 10.143 & 4.573 & 3.591 & \underline{0.297} & 20.208 & 0.715 \\
Generated-only & \underline{0.040} & 10.213 & \underline{4.629} & 3.612 & 0.294 & 20.224 & \underline{0.723} \\
Content-Matched-only & 0.057 & \textbf{10.296} & 4.620 & \underline{3.641} & \underline{0.297} & \underline{20.230} & \underline{0.723} \\
Support-only & 0.050 & 10.216 & \textbf{4.630} & \textbf{3.652} & \underline{0.297} & 20.224 & 0.716 \\
GDRM & \textbf{0.029} & \underline{10.277} & 4.621 & 3.624 & \textbf{0.298} & \textbf{20.231} & \textbf{0.728} \\
\bottomrule
\end{tabular*}
\end{table*}

\begin{table*}[tbp]
\centering
\caption{\textbf{Neutral-prompt preservation on Anime video (Wan2.1-T2V-14B).} Leakage is the mean increase in the RGB attribute logit relative to Base (lower is better). SC, DD, and Aes. denote VBench subject consistency, dynamic degree, and aesthetic quality; VQ, TA, and PC denote VideoScore2 visual quality, text alignment, and physical consistency. Each reported mean is the arithmetic mean of its three corresponding metrics. Base is shown only as a reference; \textbf{bold} and underlining indicate the best and second-best post-trained values, respectively, with ties determined at the displayed precision.}
\label{tab:video-post-neutral}
\small
\setlength{\tabcolsep}{2.6pt}
\renewcommand{\arraystretch}{1.06}
\begin{tabular*}{\linewidth}{@{\extracolsep{\fill}}lrrrrrrrrrr@{}}
\toprule
 & \multicolumn{2}{c}{Attribute} & \multicolumn{4}{c}{VBench} & \multicolumn{4}{c}{VideoScore2} \\
\cmidrule(lr){2-3}\cmidrule(lr){4-7}\cmidrule(lr){8-11}
Method & \multicolumn{2}{c}{Leakage $\downarrow$} & SC $\uparrow$ & DD $\uparrow$ & Aes. $\uparrow$ & Mean $\uparrow$ & VQ $\uparrow$ & TA $\uparrow$ & PC $\uparrow$ & Mean $\uparrow$ \\
\midrule
\rowcolor{gray!10}
Base (ref.) & \multicolumn{2}{c}{0.0000} & 0.9706 & 0.3467 & 0.6689 & 0.6621 & 4.3207 & 4.4130 & 4.1911 & 4.3083 \\
Direct Real & \multicolumn{2}{c}{0.0281} & 0.9695 & \textbf{0.3650} & \textbf{0.6626} & \textbf{0.6657} & 4.3840 & \underline{4.4630} & \textbf{4.2984} & \textbf{4.3818} \\
Generated-only & \multicolumn{2}{c}{0.1033} & 0.9729 & 0.2700 & 0.6515 & 0.6315 & 4.3683 & 4.3919 & 4.1496 & 4.3033 \\
Content-Matched-only & \multicolumn{2}{c}{0.0285} & \underline{0.9736} & \underline{0.3050} & \underline{0.6546} & 0.6444 & 4.3807 & 4.4385 & 4.2003 & 4.3398 \\
Support-only & \multicolumn{2}{c}{\textbf{0.0052}} & \underline{0.9736} & 0.2850 & 0.6526 & 0.6371 & \textbf{4.3919} & 4.3660 & 4.1834 & 4.3138 \\
GDRM & \multicolumn{2}{c}{\underline{0.0101}} & \textbf{0.9738} & \underline{0.3050} & \underline{0.6546} & \underline{0.6445} & \underline{4.3854} & \textbf{4.4665} & \underline{4.2143} & \underline{4.3554} \\
\bottomrule
\end{tabular*}
\end{table*}

\begin{table*}[tbp]
\centering
\caption{\textbf{Neutral-prompt preservation with and without TARO under two optimizer families.} VADER-based results use Wan2.1-T2V-14B, whereas Flow-GRPO-based results use Wan2.1-T2V-1.3B; comparisons are made within each panel and backbone. Metrics follow \cref{tab:video-post-neutral}. Base is shown as a reference; \textbf{bold} and underlining indicate the best and second-best post-trained values, respectively, within each panel.}
\label{tab:taro-optimizer-neutral}
\small
\setlength{\tabcolsep}{1.6pt}
\renewcommand{\arraystretch}{1.06}
\begin{tabular*}{\linewidth}{@{\extracolsep{\fill}}lrrrrrrrrrr@{}}
\toprule
\multicolumn{11}{c}{\textbf{(a) VADER-based optimization}} \\
\addlinespace[2pt]
 & \multicolumn{2}{c}{Attribute} & \multicolumn{4}{c}{VBench} & \multicolumn{4}{c}{VideoScore2} \\
\cmidrule(lr){2-3}\cmidrule(lr){4-7}\cmidrule(lr){8-11}
Method & \multicolumn{2}{c}{Leakage $\downarrow$} & SC $\uparrow$ & DD $\uparrow$ & Aes. $\uparrow$ & Mean $\uparrow$ & VQ $\uparrow$ & TA $\uparrow$ & PC $\uparrow$ & Mean $\uparrow$ \\
\midrule
Base (ref.) & \multicolumn{2}{c}{0.0000} & 0.9706 & 0.3467 & 0.6689 & 0.6621 & 4.3207 & 4.4130 & 4.1911 & 4.3083 \\
VADER & \multicolumn{2}{c}{0.0341} & \textbf{0.9742} & \textbf{0.3050} & 0.6526 & 0.6439 & 4.3217 & 4.3756 & 4.1806 & 4.2926 \\
VADER+TARO & \multicolumn{2}{c}{\textbf{0.0101}} & 0.9738 & \textbf{0.3050} & \textbf{0.6546} & \textbf{0.6445} & \textbf{4.3854} & \textbf{4.4665} & \textbf{4.2143} & \textbf{4.3554} \\
\addlinespace[3pt]
\midrule
\multicolumn{11}{c}{\textbf{(b) Flow-GRPO-based optimization}} \\
\addlinespace[2pt]
 & \multicolumn{2}{c}{Attribute} & \multicolumn{4}{c}{VBench} & \multicolumn{4}{c}{VideoScore2} \\
\cmidrule(lr){2-3}\cmidrule(lr){4-7}\cmidrule(lr){8-11}
Method & \multicolumn{2}{c}{Leakage $\downarrow$} & SC $\uparrow$ & DD $\uparrow$ & Aes. $\uparrow$ & Mean $\uparrow$ & VQ $\uparrow$ & TA $\uparrow$ & PC $\uparrow$ & Mean $\uparrow$ \\
\midrule
Base (ref.) & \multicolumn{2}{c}{0.0000} & 0.9742 & 0.3250 & 0.6568 & 0.6520 & 4.3037 & 4.2798 & 4.1059 & 4.2298 \\
Flow-GRPO & \multicolumn{2}{c}{0.1425} & \textbf{0.9728} & 0.3700 & 0.6402 & 0.6610 & 4.2099 & 4.2785 & 4.0987 & 4.1957 \\
Flow-GRPO+TARO & \multicolumn{2}{c}{\textbf{0.0631}} & 0.9668 & \textbf{0.4400} & \textbf{0.6437} & \textbf{0.6835} & \textbf{4.3409} & \textbf{4.3494} & \textbf{4.1422} & \textbf{4.2775} \\
\bottomrule
\end{tabular*}
\end{table*}

\subsection{SFT Comparison}

\Cref{tab:sft-comparison} compares positive-only LoRA fine-tuning with reward-guided post-training. SFT improves the global style attributes Oil Painting, Vintage Faded, and Anime, but provides limited benefit for the finer-grained Muscular attribute. Reward-guided methods generally yield larger and more consistent target gains. Among these methods, GDRM provides the strongest image-attribute enhancement and the highest Anime mean and median, while Direct Real remains competitive on Anime P20.

\begin{table*}[tbp]
\centering
\caption{\textbf{Positive-only SFT versus reward-guided post-training.} We report RGB target-strength percentiles for three image attributes and Anime video generated with Wan2.1-T2V-14B. SFT applies LoRA updates using only real positive examples, whereas Direct Real trains a reward model on real positive and negative examples with their original labels. Base is shown only as a reference; \textbf{bold} and underlining indicate the best and second-best reward-guided results, respectively, for each metric.}
\label{tab:sft-comparison}
\small
\setlength{\tabcolsep}{1.6pt}
\renewcommand{\arraystretch}{1.08}
\begin{tabular*}{\linewidth}{@{\extracolsep{\fill}}lrrrrrrrrrrrr@{}}
\toprule
& \multicolumn{3}{c}{\textbf{Muscular}}
& \multicolumn{3}{c}{\textbf{Oil Painting}}
& \multicolumn{3}{c}{\textbf{Vintage Faded}}
& \multicolumn{3}{c}{\textbf{Anime (video)}} \\
\cmidrule(lr){2-4}
\cmidrule(lr){5-7}
\cmidrule(lr){8-10}
\cmidrule(lr){11-13}
Method
& Median $\uparrow$ & Mean $\uparrow$ & P20 $\uparrow$
& Median $\uparrow$ & Mean $\uparrow$ & P20 $\uparrow$
& Median $\uparrow$ & Mean $\uparrow$ & P20 $\uparrow$
& Median $\uparrow$ & Mean $\uparrow$ & P20 $\uparrow$ \\
\midrule
\rowcolor{gray!10}
Base (ref.)
& 48.184 & 51.838 & 24.835
& 48.018 & 51.282 & 26.436
& 49.264 & 50.733 & 22.606
& 60.778 & 59.263 & 43.430 \\

SFT
& 47.776 & 50.460 & 24.319
& 64.091 & 60.614 & 33.835
& 57.300 & 56.889 & 26.660
& 65.087 & 66.774 & 50.822 \\

Direct Real
& \underline{57.867} & \underline{57.341} & \underline{29.196}
& \underline{80.128} & \underline{74.455} & \underline{55.177}
& \underline{77.205} & \underline{72.862} & \underline{54.527}
& \underline{68.066} & \underline{68.579} & \textbf{55.417} \\

\addlinespace[1pt]
\cmidrule(lr){1-13}
GDRM
& \textbf{65.802} & \textbf{62.862} & \textbf{36.896}
& \textbf{82.459} & \textbf{76.102} & \textbf{58.436}
& \textbf{80.903} & \textbf{74.425} & \textbf{54.624}
& \textbf{69.045} & \textbf{68.722} & \underline{54.267} \\
\bottomrule
\end{tabular*}
\end{table*}

\subsection{Qualitative TARO Comparison}

We also consider qualitative differences between raw-reward maximization and TARO for Anime video, focusing on the preservation of salient subject appearance and scene details during target-attribute enhancement.

\IfFileExists{supplementary/figures/TABO-vis-compare.pdf}{%
\begin{figure*}[t]
    \centering
    \includegraphics[width=0.92\textwidth,height=0.78\textheight,keepaspectratio]{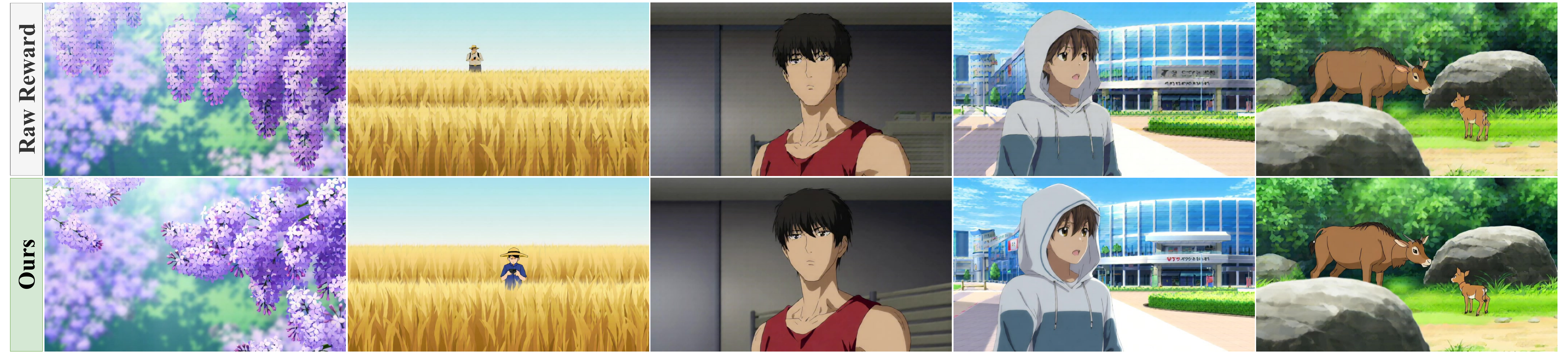}
    \caption{Qualitative comparison between raw-reward maximization and TARO on 2D Japanese Anime video. Raw-reward maximization shows more pronounced visual degradation, whereas TARO better preserves subject appearance and scene content. Top: raw reward; bottom: TARO.}
    \label{fig:bounded-vs-raw-rm-anime}
\end{figure*}
}{}

\section{Evaluator Validation and Generalization}
\label{app:validation_generalization}

\subsection{RGB-Evaluator Validation}
\label{app:evaluator_validation}

For images, we report generated-pair preference and AUROC on the real validation split. The AUROC compares evaluator scores for positive samples against all real negative samples. For Anime video, we report AUROC, AUPRC, and balanced accuracy on the real validation split.

\par\medskip
\noindent\begin{minipage}{\linewidth}
    \centering
    \small
    \captionsetup{hypcap=false}
    \captionof{table}{Image RGB-evaluator validation results.}
    \label{tab:image_rgb_validation}
    \setlength{\tabcolsep}{4pt}
    \begin{tabular}{@{}lcc@{}}
        \toprule
        Tag & \shortstack{Generated-pair\\preference} & \shortstack{Real-validation\\AUROC} \\
        \midrule
        Vintage Faded & 0.9705 & 0.955372 \\
        Oil Painting & 0.9580 & 0.998553 \\
        Muscular & 0.9715 & 0.987122 \\
        \bottomrule
    \end{tabular}
\end{minipage}

\par\medskip
\noindent\begin{minipage}{\linewidth}
    \centering
    \small
    \captionsetup{hypcap=false}
    \captionof{table}{Video RGB-evaluator validation results for Anime.}
    \label{tab:video_rgb_validation}
    \begin{tabular}{lc}
        \toprule
        Metric & Validation value \\
        \midrule
        AUROC & 0.997276 \\
        AUPRC & 0.990599 \\
        Balanced accuracy & 0.985641 \\
        \bottomrule
    \end{tabular}
\end{minipage}

\begin{figure*}[t]
    \centering
    \includegraphics[width=0.88\textwidth]{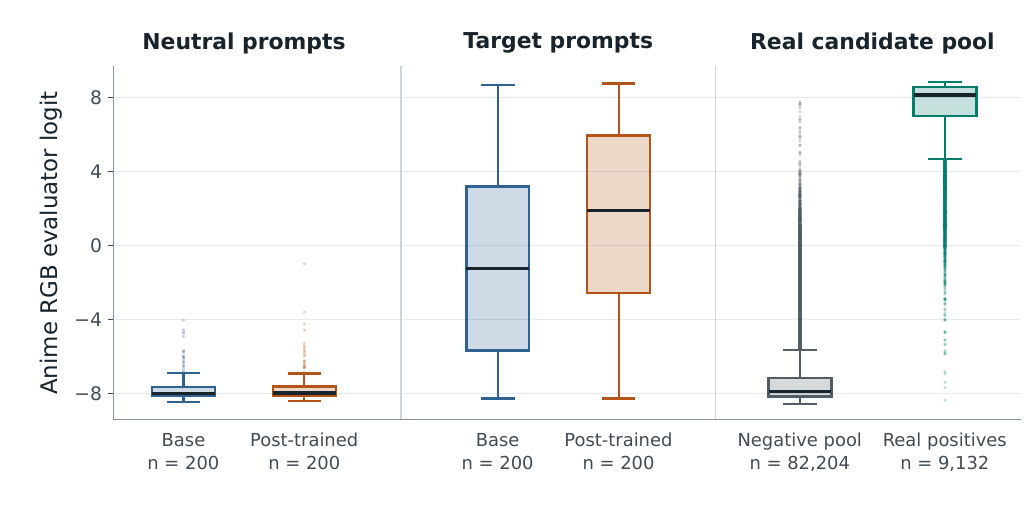}
    \caption{\textbf{Anime RGB-evaluator logit distributions.} Real-positive and negative videos from the real candidate pool (right), and Base versus post-trained Wan2.1-T2V-14B generations on 200 neutral and 200 Anime-targeted evaluation prompts (left and middle).}
    \label{fig:anime_rgb_evaluator_distribution}
\end{figure*}

\paragraph{Anime evaluator score distributions.}
As shown in \cref{fig:anime_rgb_evaluator_distribution}, the RGB evaluator clearly separates real Anime-positive and negative videos. On 200 evaluation prompts per condition, post-trained Wan2.1 retains similarly low Anime logits to Base for neutral prompts, while shifting the target-conditioned logits upward. This supports stronger Anime-style instruction following without noticeable attribute leakage into neutral generations.

\subsection{Extended VBench and VideoReward Diagnostics}
\label{app:video_post_target_full}

\Cref{tab:video-post-target-full} reports additional post-target and post-neutral diagnostics for Anime on Wan2.1-T2V-14B. This table averages six VBench components, whereas the primary video evaluation uses a three-component VBench mean. It also reports VideoReward (visual quality, motion quality, and text alignment), which is distinct from VideoScore2 (visual quality, text alignment, and physical consistency).

\begin{table*}[!tbp]
\centering
\caption{\textbf{Extended post-target and post-neutral diagnostics for Anime video (Wan2.1-T2V-14B).} The six VBench components are subject consistency (SC), background consistency (BC), motion smoothness (MS), dynamic degree (DD), aesthetic quality (Aes.), and imaging quality (IQ). VideoReward reports visual quality (VQ), motion quality (MQ), and text alignment (TA). These diagnostics complement the three-component VBench and VideoScore2 metrics reported in the primary evaluation tables. Base is shown as a reference; bold and underlining mark the two highest post-trained values.}
\label{tab:video-post-target-full}
\small
\setlength{\tabcolsep}{1.6pt}
\renewcommand{\arraystretch}{1.06}
\begin{tabular*}{\linewidth}{@{\extracolsep{\fill}}lrrrrrrrrrrr@{}}
\toprule
\multicolumn{12}{c}{\textbf{\textit{(a) Post-target enhancement}}} \\
\addlinespace[2pt]
 & \multicolumn{7}{c}{VBench} & \multicolumn{4}{c}{VideoReward} \\
\cmidrule(lr){2-8}\cmidrule(lr){9-12}
Method & SC $\uparrow$ & BC $\uparrow$ & MS $\uparrow$ & DD $\uparrow$ & Aes. $\uparrow$ & IQ $\uparrow$ & Mean $\uparrow$ & VQ $\uparrow$ & MQ $\uparrow$ & TA $\uparrow$ & Mean $\uparrow$ \\
\midrule
\rowcolor{gray!10}
Base (ref.) & 0.9725 & 0.9797 & 0.9932 & 0.2117 & 0.6655 & 0.6888 & 0.7519 & -0.0836 & -0.4643 & 0.1657 & -0.1274 \\
Direct Real & 0.9717 & 0.9763 & 0.9929 & 0.1350 & 0.6362 & 0.6824 & 0.7324 & -0.1735 & \underline{-0.5249} & 0.1437 & -0.1849 \\
Generated-only & 0.9684 & 0.9763 & \underline{0.9941} & 0.1150 & 0.6288 & 0.6877 & 0.7284 & -0.2787 & \textbf{-0.4970} & 0.1179 & -0.2193 \\
Content-Matched-only & \underline{0.9770} & 0.9803 & 0.9940 & \underline{0.1400} & \textbf{0.6444} & \textbf{0.7218} & \underline{0.7429} & \textbf{-0.1349} & -0.5329 & \textbf{0.1712} & \textbf{-0.1656} \\
Support-only & 0.9766 & \textbf{0.9808} & \textbf{0.9944} & 0.0900 & 0.6356 & 0.7113 & 0.7314 & -0.2136 & -0.5274 & 0.1389 & -0.2007 \\
GDRM & \textbf{0.9772} & \underline{0.9805} & \underline{0.9941} & \textbf{0.1450} & \underline{0.6434} & \underline{0.7177} & \textbf{0.7430} & \underline{-0.1575} & -0.5342 & \underline{0.1703} & \underline{-0.1738} \\
\addlinespace[3pt]
\midrule
\addlinespace[1pt]
\multicolumn{12}{c}{\textbf{\textit{(b) Post-neutral preservation}}} \\
\addlinespace[2pt]
 & \multicolumn{7}{c}{VBench} & \multicolumn{4}{c}{VideoReward} \\
\cmidrule(lr){2-8}\cmidrule(lr){9-12}
Method & SC $\uparrow$ & BC $\uparrow$ & MS $\uparrow$ & DD $\uparrow$ & Aes. $\uparrow$ & IQ $\uparrow$ & Mean $\uparrow$ & VQ $\uparrow$ & MQ $\uparrow$ & TA $\uparrow$ & Mean $\uparrow$ \\
\midrule
\rowcolor{gray!10}
Base (ref.) & 0.9706 & 0.9785 & 0.9890 & 0.3467 & 0.6689 & 0.7025 & 0.7760 & -0.0605 & -0.3923 & 0.6391 & 0.0621 \\
Direct Real & 0.9695 & 0.9763 & 0.9887 & \textbf{0.3650} & \textbf{0.6626} & 0.7006 & \textbf{0.7771} & -0.0862 & -0.3837 & 0.6473 & 0.0591 \\
Generated-only & 0.9729 & 0.9738 & \textbf{0.9906} & 0.2700 & 0.6515 & 0.7094 & 0.7614 & -0.1643 & -0.4098 & \underline{0.6768} & 0.0342 \\
Content-Matched-only & \underline{0.9736} & \underline{0.9769} & \underline{0.9903} & \underline{0.3050} & \underline{0.6546} & \underline{0.7105} & 0.7685 & -0.0709 & -0.3761 & 0.6494 & 0.0675 \\
Support-only & \underline{0.9736} & \underline{0.9769} & \underline{0.9903} & 0.2850 & 0.6526 & 0.7094 & 0.7646 & \textbf{-0.0642} & \underline{-0.3695} & 0.6582 & \underline{0.0748} \\
GDRM & \textbf{0.9738} & \textbf{0.9773} & 0.9902 & \underline{0.3050} & \underline{0.6546} & \textbf{0.7113} & \underline{0.7687} & \underline{-0.0651} & \textbf{-0.3672} & \textbf{0.6825} & \textbf{0.0834} \\
\bottomrule
\end{tabular*}
\end{table*}

\subsection{Generalization to Held-Out VBench Prompts}
\label{app:anime_heldout_vbench}

We assess prompt generalization using 769 unique VBench-derived prompts. Their normalized text does not overlap with the training prompts, and they are disjoint from the prompts used for checkpoint selection. Base and post-trained Wan2.1 generations use identical Anime-conditioned prompts and sampling settings. We measure Anime strength with the independent RGB evaluator and report 14 VBench dimensions, excluding Appearance Style and Temporal Flickering.

As shown in \cref{tab:anime-heldout-vbench}, the mean Anime RGB logit improves from $-4.9219$ to $-3.0821$ (a gain of $1.8398$), indicating that the target enhancement transfers to held-out prompt texts. Changes in general video metrics are mixed but modest for most dimensions: subject consistency, background consistency, and motion smoothness change by $-0.0025$, $-0.0044$, and approximately zero, respectively. Imaging quality, human action, scene, and spatial relationship improve, while dynamic degree and object class show the largest decreases ($-0.0278$ and $-0.0222$). These results indicate that targeted Anime enhancement generalizes to unseen prompt texts, with some dimension-specific trade-offs in general video performance.

\begin{table*}[!tbp]
    \centering
    \small
    \caption{\textbf{Target enhancement and general capability preservation on held-out VBench-derived prompts.} RGB logit measures anime attribute strength; the remaining rows report the 14 VBench dimensions on a 0--1 scale. $\Delta$ is the change relative to Base. VBench aggregation uses dimension-specific prompt subsets and valid samples, with imaging quality normalized by 100. All values are rounded to four decimal places;
$\Delta$ is computed using the unrounded scores.}
    \label{tab:anime-heldout-vbench}
    \setlength{\tabcolsep}{8pt}
    \renewcommand{\arraystretch}{1.08}
    \begin{tabular}{@{}lrrr@{}}
        \toprule
        Metric & Base & Ours & $\Delta$ \\
        \midrule
        \multicolumn{4}{c}{\textit{Target attribute}} \\
        \addlinespace[2pt]
        \rowcolor{gray!10}
        RGB logit (mean) & -4.9219 & \textbf{-3.0821} & \textbf{+1.8398} \\
        \midrule
        \multicolumn{4}{c}{\textit{Visual and temporal properties}} \\
        \addlinespace[2pt]
        Aesthetic quality & 0.6825 & 0.6752 & -0.0074 \\
        Imaging quality & 0.6848 & 0.6972 & +0.0124 \\
        Subject consistency & 0.9727 & 0.9703 & -0.0025 \\
        Background consistency & 0.9906 & 0.9862 & -0.0044 \\
        Motion smoothness & 0.9890 & 0.9890 & -0.0000 \\
        Dynamic degree & 0.4306 & 0.4028 & -0.0278 \\
        Temporal style & 0.2288 & 0.2239 & -0.0049 \\
        \midrule
        \multicolumn{4}{c}{\textit{Content and prompt alignment}} \\
        \addlinespace[2pt]
        Object class & 0.8465 & 0.8244 & -0.0222 \\
        Multiple objects & 0.6860 & 0.6814 & -0.0046 \\
        Human action & 0.8700 & 0.9200 & +0.0500 \\
        Color & 0.9553 & 0.9397 & -0.0157 \\
        Scene & 0.3517 & 0.3714 & +0.0196 \\
        Spatial relationship & 0.6876 & 0.6902 & +0.0026 \\
        Overall consistency & 0.2643 & 0.2639 & -0.0004 \\
        \bottomrule
    \end{tabular}
\end{table*}

\section{Ablation of Advantage-Level Adaptation for Group-Relative Policy Optimization}
\label{app:advantage_adaptation_ablation}

We isolate the effects of timestep balancing and advantage adaptation using four configurations on Wan2.1-T2V-1.3B: Base, Flow-GRPO, Flow-GRPO with timestep balancing, and the full Flow-GRPO+TARO method. All variants are evaluated on the same 200 Anime prompts.

\begin{table*}[t]
    \centering
    \small
    \caption{\textbf{Ablation of timestep balancing and advantage adaptation in Flow-GRPO.}
    Target strength is measured by the mean RGB logit and RGB-score percentiles (Mean and P20). VBench reports subject consistency (SC), dynamic degree (DD), and aesthetic quality (Aes.), together with their arithmetic mean. The final row reports the full Flow-GRPO+TARO configuration.}
    \label{tab:flowgrpo_advantage_ablation}
    \setlength{\tabcolsep}{4pt}
    \renewcommand{\arraystretch}{1.12}
    \begin{tabular*}{\textwidth}{@{\extracolsep{\fill}}p{0.34\textwidth}rrrrrrr@{}}
        \toprule
        Method
        & \shortstack{RGB logit\\mean}
        & \shortstack{Percentile\\Mean}
        & P20
        & \shortstack{VBench\\Mean}
        & SC
        & DD
        & Aes. \\
        \midrule
        Base
        & -0.1955 & 61.706 & 45.686 & 0.5849 & 0.9790 & 0.1200 & 0.6558 \\
        Flow-GRPO
        & 1.1298 & 67.227 & 51.002 & 0.5893 & 0.9775 & 0.1450 & 0.6455 \\
        Flow-GRPO + Timestep Balancing
        & 1.8959 & 70.519 & 58.712 & 0.5991 & 0.9751 & 0.1900 & 0.6323 \\
        Flow-GRPO+TARO
        & 2.2369 & 71.684 & 60.401 & 0.6286 & 0.9700 & 0.2800 & 0.6359 \\
        \bottomrule
    \end{tabular*}
\end{table*}

The results show a progressive increase in target strength as the two components are introduced. Adding timestep balancing to Flow-GRPO improves the RGB logit and target-percentile statistics; adding advantage adaptation further improves the RGB logit mean, percentile Mean/P20, and aggregate VBench mean. Relative to Base, the full method increases dynamic degree while modestly reducing subject consistency and aesthetic quality. The final row reports the complete Flow-GRPO+TARO configuration.

\section{Latent Reward-Model Logit Distributions}
\label{app:latent_rm_logit_distributions}

\begin{figure*}[t]
    \centering
    \includegraphics[width=0.62\textwidth]{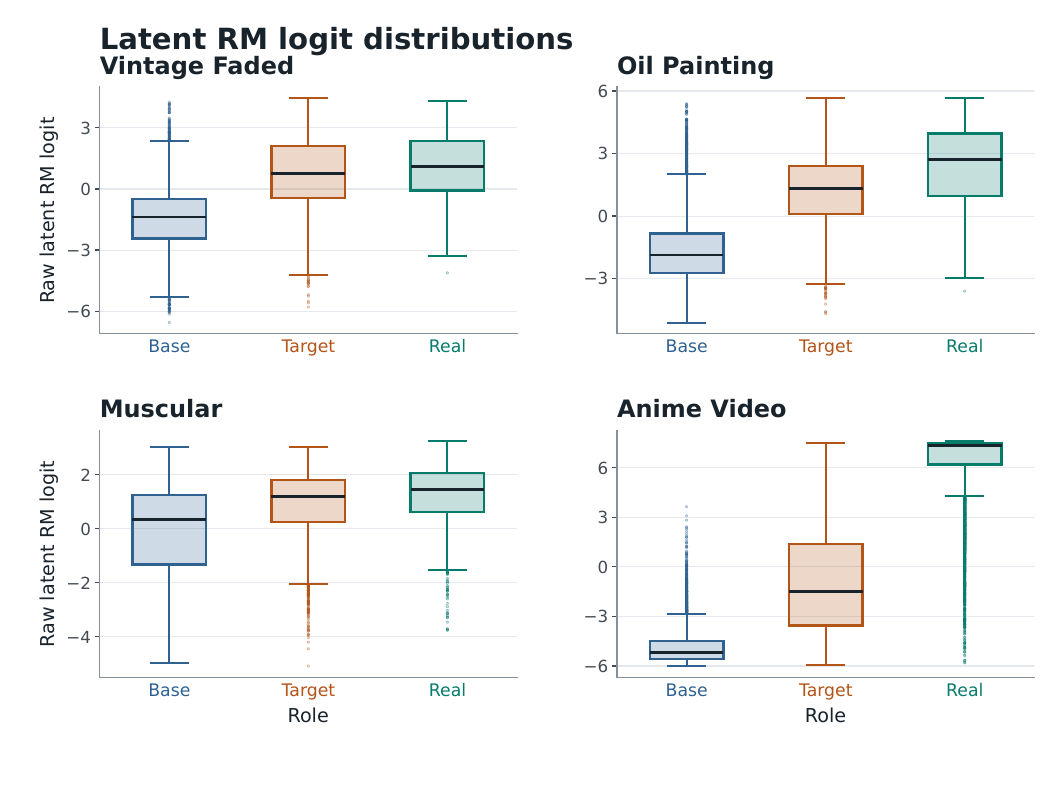}
    \caption{\textbf{Raw latent reward-model logit distributions.} Each panel compares Base anchors, Target supports, and real positive examples for one attribute. Boxes show the interquartile range, center lines mark the medians, and points beyond the whiskers denote outliers. Each panel uses an attribute-specific vertical scale; raw logit magnitudes should therefore be compared within, not across, attributes.}
    \label{fig:latent_rm_boxplots}
\end{figure*}

\Cref{fig:latent_rm_boxplots} compares raw logits from the frozen latent reward models for Base anchors, Target supports, and real positive examples. Within each attribute, Target supports have higher median logits than Base anchors, with median shifts of 2.15, 3.19, 0.87, and 3.70 for Vintage Faded, Oil Painting, Muscular, and Anime Video, respectively. This consistent shift indicates that the generator-conditioned supports capture the intended direction of attribute change.

The relationship between Target supports and real positives varies across attributes. Their interquartile ranges overlap substantially for Vintage Faded and Muscular, suggesting close alignment in these two score regimes. Oil Painting exhibits partial overlap but a higher real-positive median (2.72 versus 1.33). Anime Video shows the largest gap: the Target-support median is $-1.48$, compared with $7.34$ for real positives, while Base anchors have a median of $-5.19$. These patterns support calibrating the TARO target from real positive examples: generator supports identify a reachable attribute direction, whereas real positives provide the data-grounded reward level.

\begin{samepage}
\noindent Since the raw logit scales are attribute-dependent, comparisons should be made within each panel rather than across attributes.
\end{samepage}

\begin{figure*}[t]
    \section{Additional Qualitative Comparisons Between Base and GenTA}
    \label{app:qualitative_base_genta}
    We show matched Base and GenTA outputs for Vintage Faded, Oil Painting, Muscular, and 2D Japanese Anime. For Anime, GenTA more consistently follows the requested anime-style instruction while retaining the prompt's depicted content.\par
    \vspace{5pt}
    \centering
    \includegraphics[width=0.76\textwidth,height=0.68\textheight,keepaspectratio]{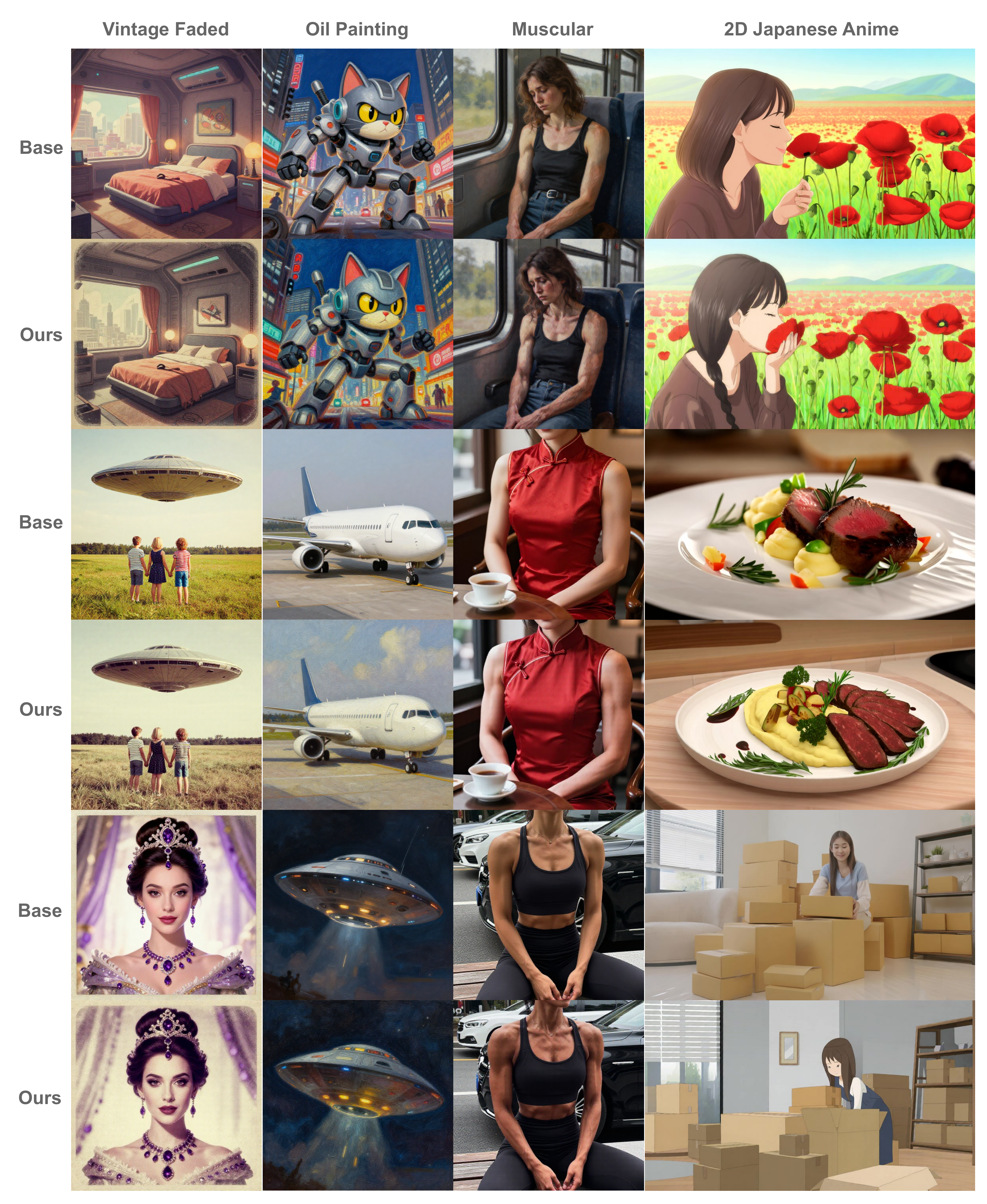}
    \caption{\textbf{Additional Base-versus-GenTA comparisons across four attributes.} The montage shows Vintage Faded, Oil Painting, Muscular, and 2D Japanese Anime examples. GenTA better follows the requested visual attributes; the Anime examples illustrate improved anime-instruction following while retaining the depicted content.}
    \label{fig:base_genta_four_attributes}
\end{figure*}

\begin{figure*}[t]
    \section{Additional Validation on Stable Diffusion 3.5}
    \label{app:stable_diffusion_35_validation}
    We further assess the cross-backbone generalization of the Oil Painting attribute using Stable Diffusion 3.5 Medium. Both Base and GenTA use 40 steps during training and inference. We qualitatively compare their outputs generated from matched prompts to examine whether the target visual style can be induced on this additional backbone.\par
    \vspace{5pt}
    \centering
    \includegraphics[width=0.88\textwidth,height=0.68\textheight,keepaspectratio]{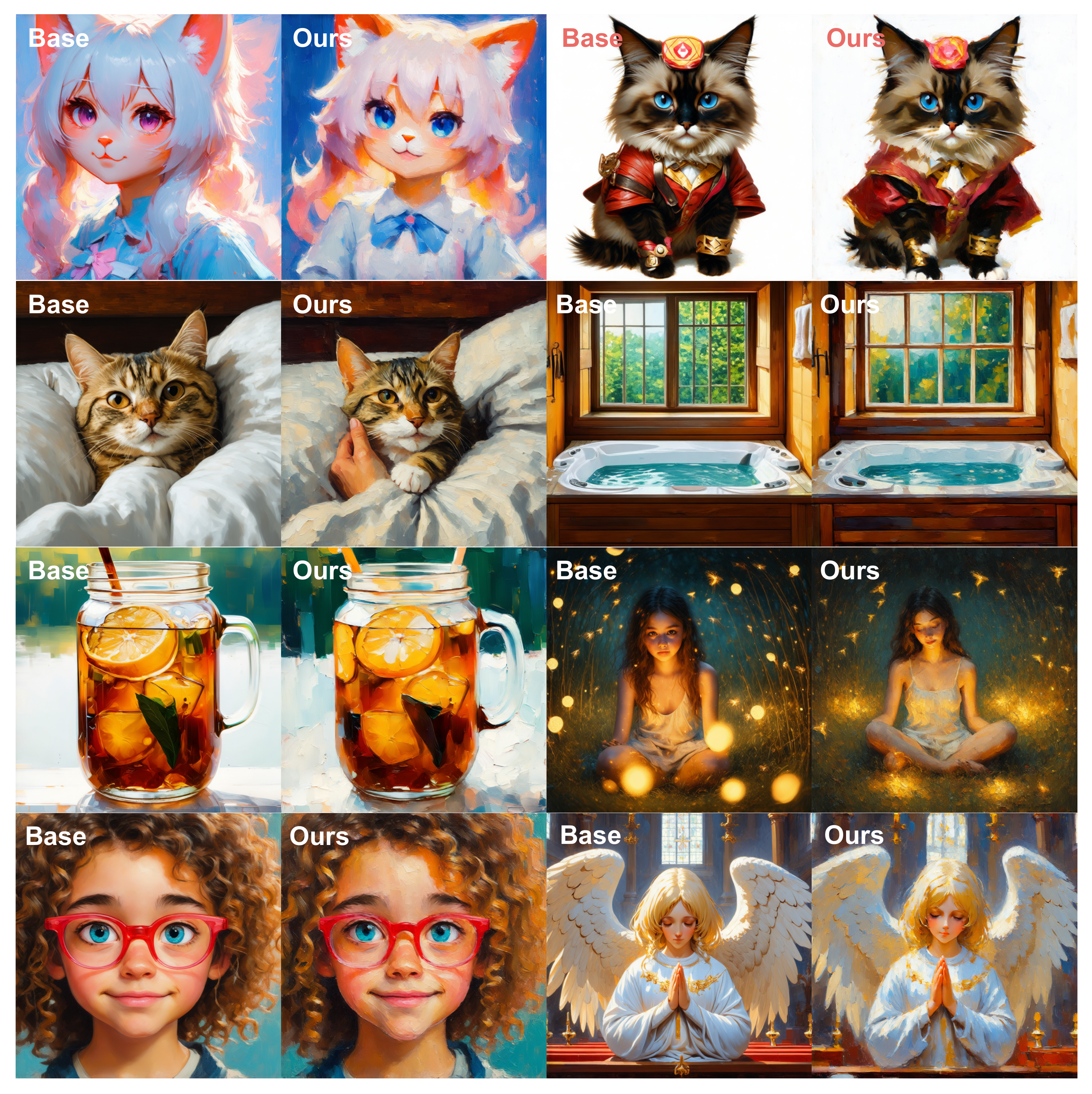}
    \caption{\textbf{Qualitative cross-backbone validation on Stable Diffusion 3.5 Medium.} Matched Base and GenTA outputs are compared for the Oil Painting attribute.}
    \label{fig:sd35_vintage_oil}
\end{figure*}

\begin{figure*}[t]
    \section{Qualitative Generalization on Held-Out VBench-Derived Prompts}
    \label{app:heldout_vbench_qualitative}
    We visualize matched Base and GenTA outputs for diverse held-out VBench-derived prompts, including human actions, object interactions, animals, and scene descriptions. Across these examples, GenTA improves the Base model's ability to follow anime-style instructions while preserving the requested subjects, actions, and scenes.\par
    \vspace{5pt}
    \centering
    \includegraphics[width=0.90\textwidth,height=0.62\textheight,keepaspectratio]{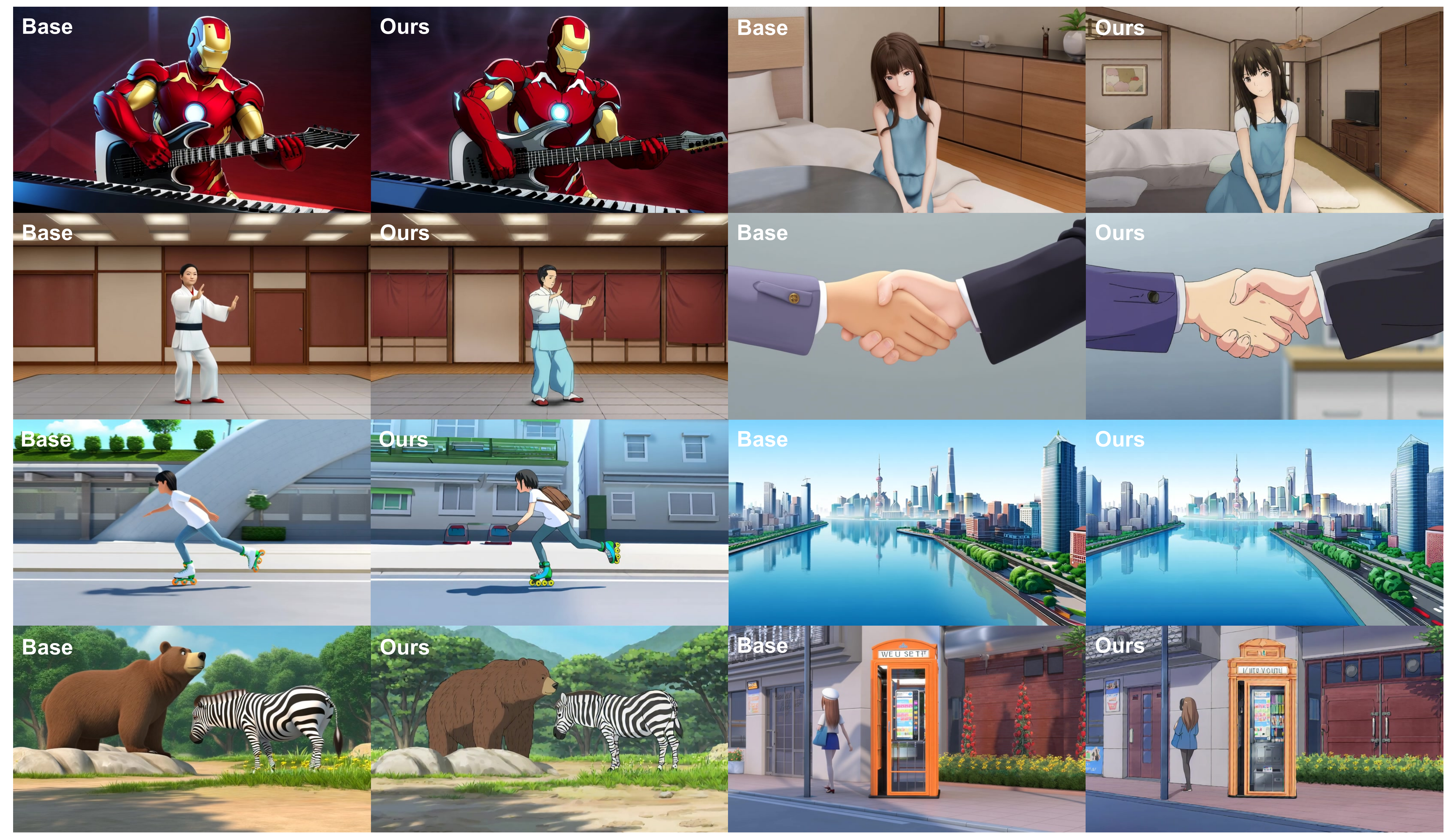}
    \caption{\textbf{Qualitative comparison on held-out VBench-derived prompts.} Across diverse prompts, GenTA more consistently follows the requested anime style while preserving the specified subjects, actions, and scenes.}
    \label{fig:heldout_vbench_qualitative}
\end{figure*}

\end{document}